%% file: main.tex
\documentclass[english]{article}
\usepackage[T1]{fontenc}
\usepackage[latin9]{inputenc}
\usepackage{babel}
\usepackage{array}
\usepackage{float}
\usepackage{multirow}
\usepackage{tablefootnote}
\usepackage{amsmath}
\usepackage{amsthm}
\usepackage{amssymb}
\usepackage{stackrel}
\usepackage{graphicx}
\usepackage[unicode=true,pdfusetitle,
 bookmarks=true,bookmarksnumbered=false,bookmarksopen=false,
 breaklinks=false,pdfborder={0 0 1},backref=false,colorlinks=false]
 {hyperref}

\makeatletter

\providecommand{\tabularnewline}{\\}
\floatstyle{ruled}
\newfloat{algorithm}{tbp}{loa}
\providecommand{\algorithmname}{Algorithm}
\floatname{algorithm}{\protect\algorithmname}

\theoremstyle{plain}
\newtheorem{thm}{\protect\theoremname}

\usepackage{algorithm,algpseudocode}
\usepackage{amsmath}
\algnewcommand\algorithmicforeach{\textbf{for each}}
\algdef{S}[FOR]{ForEach}[1]{\algorithmicforeach\ #1\ \algorithmicdo}

\usepackage{iclr2019_conference}
\iclrfinalcopy 
\renewcommand{\cite}{\citep}
\usepackage{hyperref} 

\makeatother

\providecommand{\theoremname}{Theorem}

\begin{document}
\title{Learning to Remember More \\ with Less Memorization}
\author{Hung Le, Truyen Tran and Svetha Venkatesh\\
Applied AI Institute, Deakin University, Geelong, Australia\\
\texttt{\{lethai,truyen.tran,svetha.venkatesh\}@deakin.edu.au}}
\maketitle
\begin{abstract}
\input{abs.tex} 
\end{abstract}

\section{Introduction}

\input{intro.tex}

\section{Methods}

\input{method.tex}

\section{Results}

\input{exp.tex}

\section{Related Work}

\input{related.tex}

\section{Conclusions}

\input{discuss.tex}

\bibliographystyle{iclr2019_conference}
\bibliography{}

\newpage{}

\renewcommand\thesubsection{\Alph{subsection}}

\section*{Appendix}

\input{appendix.tex}

\end{document}

%% file: abs.tex
Memory-augmented neural networks consisting of a neural controller
and an external memory have shown potentials in long-term sequential
learning. Current RAM-like memory models maintain memory accessing
every timesteps, thus they do not effectively leverage the short-term
memory held in the controller. We hypothesize that this scheme of
writing is suboptimal in memory utilization and introduces redundant
computation. To validate our hypothesis, we derive a theoretical bound
on the amount of information stored in a RAM-like system and formulate
an optimization problem that maximizes the bound. The proposed solution
dubbed Uniform Writing is proved to be optimal under the assumption
of equal timestep contributions. To relax this assumption, we introduce
modifications to the original solution, resulting in a solution termed
Cached Uniform Writing. This method aims to balance between maximizing
memorization and forgetting via overwriting mechanisms. Through an
extensive set of experiments, we empirically demonstrate the advantages
of our solutions over other recurrent architectures, claiming the
state-of-the-arts in various sequential modeling tasks. 

%% file: intro.tex
A core task in sequence learning is to capture long-term dependencies
amongst timesteps which demands memorization of distant inputs. In
recurrent neural networks (RNNs), the memorization is implicitly executed
via integrating the input history into the state of the networks.
However, learning vanilla RNNs over long distance proves to be difficult
due to the vanishing gradient problem \cite{bengio1994learning,pascanu2013difficulty}.
One alleviation is to introduce skip-connections along the execution
path, in the forms of dilated layers \cite{van2016wavenet,chang2017dilated},
attention mechanisms \cite{bahdanau2015neural,vaswani2017attention}
and external memory \cite{graves2014neural,graves2016hybrid}.

Amongst all, using external memory most resembles human cognitive
architecture where we perceive the world sequentially and make decision
by consulting our memory. Recent attempts have simulated this process
by using RAM-like memory architectures that store information into
memory slots. Reading and writing are governed by neural controllers
using attention mechanisms. These memory-augmented neural networks
(MANN) have demonstrated superior performance over recurrent networks
in various synthetic experiments \cite{graves2016hybrid} and realistic
applications \cite{le2018variational,Le:2018:DMN:3219819.3219981,W18-2606}.

Despite the promising empirical results, there is no theoretical analysis
or clear understanding on optimal operations that a memory should
have to maximize its performance. To the best of our knowledge, no
solution has been proposed to help MANNs handle ultra-long sequences
given limited memory. This scenario is practical because (i) sequences
in the real-world can be very long while the computer resources are
limited and (ii) it reflects the ability to compress in human brain
to perform life-long learning. Previous attempts such as \cite{rae2016scaling}
try to learn ultra-long sequences by expanding the memory, which is
not always feasible and do not aim to optimize the memory by some
theoretical criterion. This paper presents a new approach towards
finding optimal operations for MANNs that serve the purpose of learning
longer sequences with finite memory. 

More specifically, upon analyzing RNN and MANN operations we first
introduce a measurement on the amount of information that a MANN holds
after encoding a sequence. This metric reflects the quality of memorization
under the assumption that contributions from timesteps are equally
important. We then derive a generic solution to optimize the measurement.
We term this optimal solution as Uniform Writing (UW), and it is applicable
for any MANN due to its generality. Crucially, UW helps reduce significantly
the computation time of MANN. Third, to relax the assumption and enable
the method to work in realistic settings, we further propose Cached
Uniform Writing (CUW) as an improvement over the Uniform Writing scheme.
By combining uniform writing with local attention, CUW can learn to
discriminate timesteps while maximizing local memorization. Finally
we demonstrate that our proposed models outperform several MANNs and
other state-of-the-art methods in various synthetic and practical
sequence modeling tasks.

%% file: method.tex
\subsection{Theoretical Analysis}

Memory-augmented neural networks can be viewed as an extension of
RNNs with external memory $M$. The memory supports read and write
operations based on the output $o_{t}$ of the controller, which in
turn is a function of current timestep input $x_{t}$, previous hidden
state $h_{t-1}$ and read value $r_{t-1}$ from the memory. Let assume
we are given these operators from recent MANNs such as NTM \cite{graves2014neural}
or DNC \cite{graves2016hybrid}, represented as:

\begin{minipage}[t]{0.5\textwidth}%
\begin{equation}
r_{t}=f_{r}\left(o_{t},M_{t-1}\right)\label{eq:read_R}
\end{equation}
\end{minipage}%
\begin{minipage}[t]{0.5\textwidth}%
\begin{equation}
M_{t}=f_{w}\left(o_{t},M_{t-1}\right)\label{eq:write_v}
\end{equation}
\end{minipage}

The controller output and hidden state are updated as follows:

\begin{minipage}[t]{0.5\textwidth}%
\begin{equation}
o_{t}=f_{o}\left(h_{t-1},r_{t-1},x_{t}\right)\label{eq:update_o}
\end{equation}
\end{minipage}%
\begin{minipage}[t]{0.5\textwidth}%
\begin{equation}
h_{t}=f_{h}\left(h_{t-1},r_{t-1},x_{t}\right)\label{eq:control_h}
\end{equation}
\end{minipage}

Here, $f_{o}$ and $f_{h}$ are often implemented as RNNs while $f_{r}$
and $f_{w}$ are designed specifically for different memory types. 

Current MANNs only support regular writing by applying Eq. (\ref{eq:write_v})
every timestep. In effect, regular writing ignores the accumulated
short-term memory stored in the controller hidden states which may
well-capture the recent subsequence. We argue that the controller
does not need to write to memory continuously as its hidden state
also supports memorizing. Another problem of regular writing is time
complexity. As the memory access is very expensive, reading/writing
at every timestep makes MANNs much slower than RNNs. This motivates
a irregular writing strategy to utilize the memorization capacity
of the controller and and consequently, speed up the model. In the
next sections, we first define a metric to measure the memorization
performance of RNNs, as well as MANNs. Then, we solve the problem
of finding the best irregular writing that optimizes the metric. 

\subsubsection{Memory analysis of RNNs}

We first define the ability to ``remember'' of recurrent neural
networks, which is closely related to the vanishing/exploding gradient
problem \cite{pascanu2013difficulty}. In RNNs, the state transition
$h_{t}=\phi\left(h_{t-1},x_{t}\right)$ contains contributions from
not only $x_{t}$, but also previous timesteps $x_{i<t}$ embedded
in $h_{t-1}$. Thus, $h_{t}$ can be considered as a function of timestep
inputs, i.e, $h_{t}=f\left(x_{1},x_{2},...,x_{t}\right)$. One way
to measure how much an input $x_{i}$ contributes to the value of
$h_{t}$ is to calculate the norm of the gradient $\left\Vert \frac{\partial h_{t}}{\partial x_{i}}\right\Vert $.
If the norm equals zero, $h_{t}$ is constant w.r.t $x_{i}$, that
is, $h_{t}$ does not ``remember'' $x_{i}$. As a bigger $\left\Vert \frac{\partial h_{t}}{\partial x_{i}}\right\Vert $
implies more influence of $x_{i}$ on $h_{t}$, we propose using $\left\Vert \frac{\partial h_{t}}{\partial x_{i}}\right\Vert $
to measure the contribution of the $i$-th input to the $t$-th hidden
state. Let $c_{i,t}$ denotes this term, we can show that in the case
of common RNNs, $\lambda_{c}c_{i,t}\geq c_{i-1,t}$ with some $\lambda_{c}\in\mathbb{R^{+}}$(see
Appendix \ref{subsec:Derivation-on-theRNN} - \ref{subsec:Derivation-on-theLSTM}
for proof). This means further to the past, the contribution decays
(when $\lambda_{c}<1$) or grows (when $\lambda_{c}>1$) with the
rate of at least $\lambda_{c}$ .We can measure the average amount
of contributions across $T$ timesteps as follows (see Appendix \ref{subsec:Proof-of-theorem-1}
for proof):
\begin{thm}
\label{thm:The-average-amount} There exists $\lambda\in\mathbb{R^{+}}$such
that the average contribution of a sequence of length T with respect
to a RNN can be quantified as the following:

\begin{equation}
I_{\lambda}=\frac{\stackrel[t=1]{T}{\sum}c_{t,T}}{T}=c_{T,T}\frac{\stackrel[t=1]{T}{\sum}\lambda^{T-t}}{T}\label{eq:d0}
\end{equation}
\end{thm}

If $\lambda<1$, $\lambda^{T-t}\rightarrow0$ as $T-t\rightarrow\infty$.
This is closely related to vanishing gradient problem. LSTM is known
to ``remember'' long sequences better than RNN by using extra memory
gating mechanisms, which help $\lambda$ to get closer to 1. If $\lambda>1$,
the system may be unstable and suffer from the exploding gradient
problem. 

\begin{figure}
\begin{centering}
\includegraphics[width=0.75\textwidth]{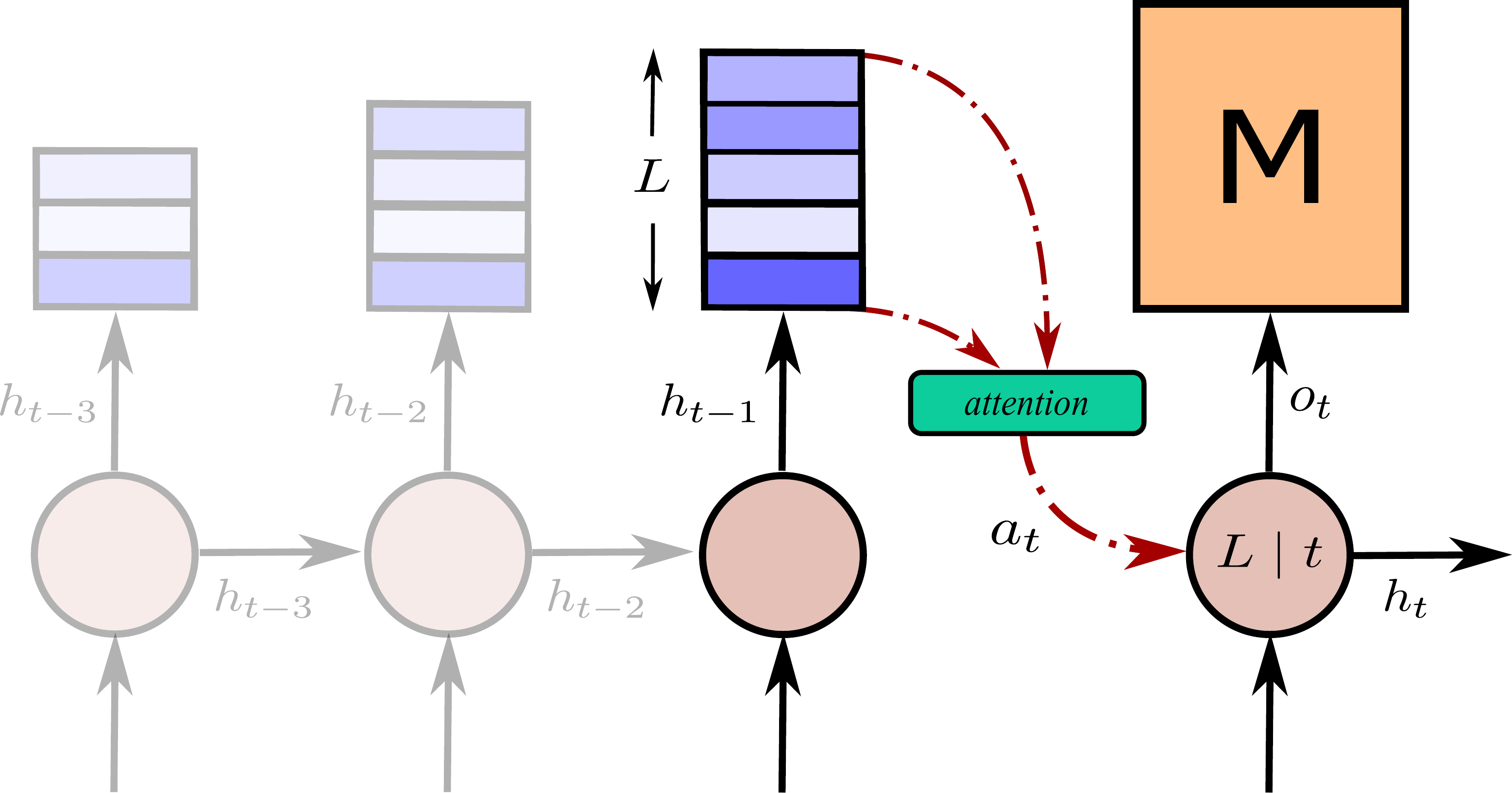}
\par\end{centering}
\caption{Writing mechanism in Cached Uniform Writing. During non-writing intervals,
the controller hidden states are pushed into the cache. When the writing
time comes, the controller attends to the cache, chooses suitable
states and accesses the memory. The cache is then emptied. \label{fig:Writing-mechanism-in}}
\end{figure}

\subsubsection{Memory analysis of MANNs}

In slot-based MANNs, memory $M$ is a set of $D$ memory slots. A
write at step $t$ can be represented by the controller's hidden state
$h_{t}$, which accumulates inputs over several timesteps (i.e., $x_{1}$,
...,$x_{t}$). If another write happens at step $t+k$, the state
$h_{t+k}$'s information containing timesteps $x_{t+1}$, ...,$x_{t+k}$
is stored in the memory ($h_{t+k}$ may involves timesteps further
to the past, yet they are already stored in the previous write and
can be ignored). During writing, overwriting may happen, replacing
an old write with a new one. Thus after all, $D$ memory slots associate
with $D$ chosen writes of the controller. From these observations,
we can generalize Theorem \ref{thm:The-average-amount} to the case
of MANNs having $D$ memory slots (see Appendix \ref{subsec:Proof-of-theorem-2}
for proof).
\begin{thm}
\label{thm:Under-the-assumption}With any $D$ chosen writes at timesteps
$1\leq$$K_{1}$\textup{$<$} $K_{2}$$<$ ...$<$$K_{D}$$<T$, there
exist $\lambda,C\in\mathbb{R^{+}}$such that the lower bound on the
average contribution of a sequence of length $T$ with respect to
a MANN having $D$ memory slots can be quantified as the following:

\begin{align}
I_{\lambda} & =C\frac{\stackrel[t=1]{K_{1}}{\sum}\lambda^{K_{1}-t}+\stackrel[t=K_{1}+1]{K_{2}}{\sum}\lambda^{K_{2}-t}+...+\stackrel[t=K_{D-1}+1]{K_{D}}{\sum}\lambda^{K_{D}-t}+\stackrel[t=K_{D}+1]{T}{\sum}\lambda^{T-t}}{T}\nonumber \\
 & =\frac{C}{T}\stackrel[i=1]{D+1}{\sum}\stackrel[j=0]{l_{i}-1}{\sum}\lambda^{j}=\frac{C}{T}\stackrel[i=1]{D+1}{\sum}f_{\lambda}(l_{i})
\end{align}
where $l_{i}=\begin{cases}
K_{1} & ;i=1\\
K_{i}-K_{i-1} & ;D\geq i>1\\
T-K_{D} & ;i=D+1
\end{cases}$, $f_{\lambda}\left(x\right)=\begin{cases}
\frac{1-\lambda^{x}}{1-\lambda} & \lambda\neq1\\
x & \lambda=1
\end{cases}$, $\forall x\in\mathbb{R}^{+}$.
\end{thm}

If $\lambda\leq1$, we want to maximize $I_{\lambda}$ to keep the
information from vanishing. On the contrary, if $\lambda>1$, we may
want to minimize $I_{\lambda}$ to prevent the information explosion.
As both scenarios share the same solution (see Appendix \ref{subsec:Proof-of-theorem}),
thereafter we assume that $\lambda\leq1$ holds for other analyses.
By taking average over $T$, we are making an assumption that all
timesteps are equally important. This helps simplify the measurement
as $I_{\lambda}$ is independent of the specific position of writing.
Rather, it is a function of the interval lengths between the writes.
This turns out to be an optimization problem whose solution is stated
in the following theorem.
\begin{thm}
\label{thm:Given-the-number}Given $D$ memory slots, a sequence with
length $T$, a decay rate $0<\lambda\leq1$, then the optimal intervals
$\left\{ l_{i}\in\mathbb{R^{+}}\right\} _{i=1}^{D+1}$ satisfying
$T=\stackrel[i=1]{D+1}{\sum}l_{i}$ such that the lower bound on the
average contribution $I_{\lambda}=\frac{C}{T}\stackrel[i=1]{D+1}{\sum}f_{\lambda}(l_{i})$
is maximized are the following:

\begin{equation}
l_{1}=l_{2}=...=l_{D+1}=\frac{T}{D+1}
\end{equation}
\end{thm}

We name the optimal solution as Uniform Writing (UW) and refer to
the term $\frac{T}{D+1}$ and $\frac{D+1}{T}$ as the \textit{optimal
interval} and the\textit{ compression ratio}, respectively. The proof
is given in Appendix \ref{subsec:Proof-of-theorem}.

\begin{algorithm}[t]
\begin{algorithmic}[1]
\Require{a sequence $x=\left\{ x_{t}\right\} _{t=1}^{T}$, a cache $C$ sized $L$, a memory sized $D$}.
\For{$t=1,T$}
\State{$C$.append($h_{t-1}$)}
\If{$t \bmod L==0$}
\State{Use Eq.($\ref{eq:att}$) to calculate $a_t$}
\State{Execute Eq.($\ref{eq:update_o}$): $o_{t}=f_{o}\left(a_t,r_{t-1},x_{t}\right)$}
\State{Execute Eq.($\ref{eq:control_h}$): $h_{t}=f_{h}\left(a_t,r_{t-1},x_{t}\right)$}
\State{Update the memory using Eq.($\ref{eq:write_v}$)}
\State{Read $r_t$ from the memory using Eq.($\ref{eq:read_R}$)}
\State{$C$.clear()}
\Else
\State{Update the controller using Eq.($\ref{eq:control_h}$): $h_{t}=f_{h}\left(h_{t-1},r_{t-1},x_{t}\right)$}
\State{Assign $r_t=r_{t-1}$}
\EndIf
\EndFor
\end{algorithmic} 

\caption{Cached Uniform Writing\label{alg:Cached-Uniform-Writing}}
\end{algorithm}

\subsection{Proposed Models}

Uniform writing can apply to any MANNs that support writing operations.
Since the writing intervals are discrete, i.e., $l_{i}\in\mathbb{N^{+}}$,
UW is implemented as the following:

\begin{equation}
M_{t}=\begin{cases}
f_{w}\left(o_{t},M_{t-1}\right) & if\:t=\left\lfloor \frac{T}{D+1}\right\rfloor k,k\in\mathbb{N^{+}}\\
M_{t-1} & otherwise
\end{cases}\label{eq:uw}
\end{equation}

By following Eq. (\ref{eq:uw}), the write intervals are close to
the optimal interval defined in Theorem \ref{thm:Given-the-number}
and approximately maximize the average contribution. This writing
policy works well if timesteps are equally important and the task
is to remember all of them to produce outputs (i.e., in copy task).
However, in reality, timesteps are not created equal and a good model
may need to ignore unimportant or noisy timesteps. That is why overwriting
in MANN can be necessary. In the next section, we propose a method
that tries to balance between following the optimal strategy and employing
overwriting mechanism as in current MANNs. 

\subsubsection{Local optimal design}

To relax the assumptions of Theorem \ref{thm:Given-the-number}, we
propose two improvements of the\textit{ }Uniform Writing (UW) strategy.
First, the intervals between writes are equal with length $L$ ($1\leq L\leq\left\lfloor \frac{T}{D+1}\right\rfloor $).
If $L=1$, the strategy becomes regular writing and if $L=\left\lfloor \frac{T}{D+1}\right\rfloor $,
it becomes uniform writing. This ensures that after $\left\lfloor \frac{T}{L}\right\rfloor $
writes, all memory slots should be filled and the model has to learn
to overwrite. Meanwhile, the average kept information is still locally
maximized every $L*D$ timesteps.

Second, we introduce a cache of size $L$ to store the hidden states
of the controller during a write interval. Instead of using the hidden
state at the writing timestep to update the memory, we perform an
attention over the cache to choose the best representative hidden
state. The model will learn to assign attention weights to the elements
in the cache. This mechanism helps the model consider the importance
of each timestep input in the local interval and thus relax the equal
contribution assumption of Theorem \ref{thm:Given-the-number}. We
name the writing strategy that uses the two mentioned-above improvements
as Cached Uniform Writing (CUW). An illustration of the writing mechanism
is depicted in Fig. \ref{fig:Writing-mechanism-in}.

\subsubsection{Local memory-augmented attention unit}

In this subsection, we provide details of the attention mechanism
used in our CUW. To be specific, the best representative hidden state
$a_{t}$ is computed as follows:

\noindent\begin{minipage}[t]{0.61\columnwidth}%
\begin{equation}
\alpha_{tj}=softmax\left(v^{T}\tanh\left(Wh_{t-1}+Ud_{j}+Vr_{t-1}\right)\right)
\end{equation}
\end{minipage}%
\begin{minipage}[t]{0.39\columnwidth}%
\begin{equation}
a_{t}=\stackrel[j=1]{L}{\sum}\alpha_{tj}d_{j}\label{eq:att}
\end{equation}
\end{minipage}

where $\alpha_{tj}$ is the attention score between the $t$-th writing
step and the $j$-th element in the cache; $W$, $U$, $V$ and $v$
are parameters; $h$ and $r$ are the hidden state of the controller
and the read-out (Eq. (\ref{eq:read_R})), respectively; $d_{j}$
is the cache element and can be implemented as the controller's hidden
state ($d_{j}=h_{t-1-L+j}$). 

The vector $a_{t}$ will be used to replace the previous hidden state
in updating the controller and memory. The whole process of performing
CUW is summarized in Algo. \ref{alg:Cached-Uniform-Writing}.

%% file: exp.tex
\begin{figure}
\begin{centering}
\includegraphics[width=1\linewidth]{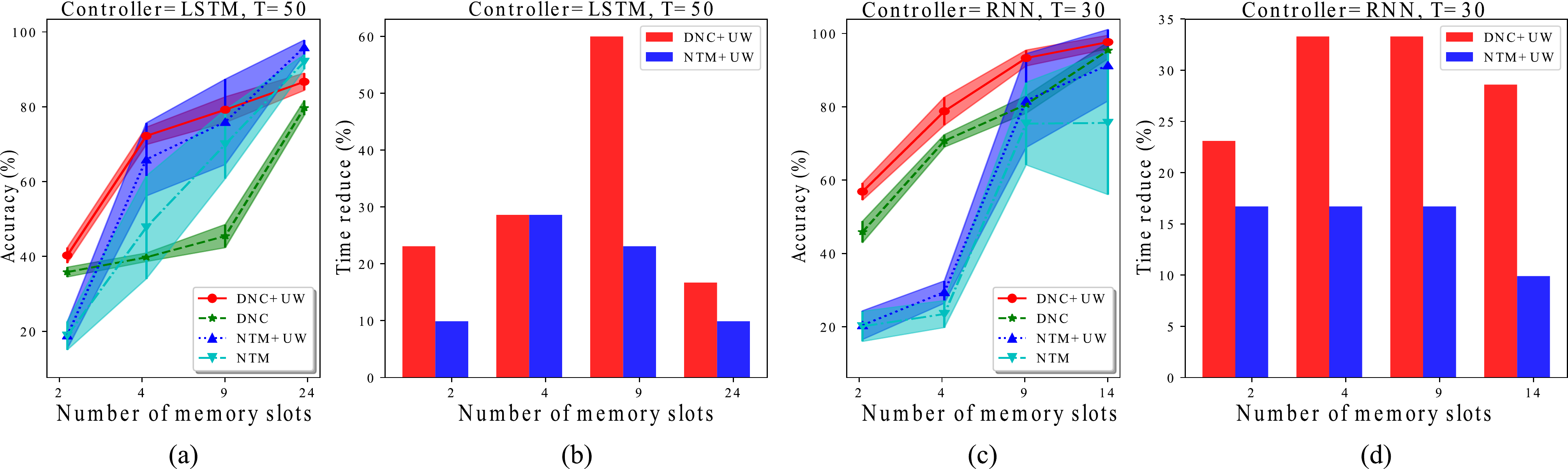}
\par\end{centering}
\caption{The accuracy (\%) and computation time reduction (\%) with different
memory types and number of memory slots. The controllers/sequence
lengths/memory sizes are chosen as LSTM/50/$\left\{ 2,4,9,24\right\} $
(a\&b) and RNN/30/$\left\{ 2,4,9,14\right\} $ (c\&d), respectively.
\label{fig:The-test-accuracy}}
\end{figure}

\subsection{An Ablation Study: Memory-augmented Neural Networks with and without
Uniform Writing}

In this section, we study the impact of uniform writing on MANNs under
various circumstances (different controller types, memory types and
number of memory slots). We restrict the memorization problem to the
double task in which the models must reconstruct a sequence of integers
sampled uniformly from range $\left[1,10\right]$ twice. We cast this
problem to a sequence to sequence problem with 10 possible outputs
per decoding step. The training stops after 10,000 iterations of batch
size 64. We choose DNC\footnote{Our reimplementation based on https://github.com/deepmind/dnc}
and NTM\footnote{https://github.com/MarkPKCollier/NeuralTuringMachine}
as the two MANNs in the experiment. The recurrent controllers can
be RNN or LSTM. With LSTM controller, the sequence length is set to
50. We choose sequence length of 30 to make it easier for the RNN
controller to learn the task. The number of memory slots $D$ is chosen
from the set $\left\{ 2,4,9,24\right\} $ and $\left\{ 2,4,9,14\right\} $
for LSTM and RNN controllers, respectively. More memory slots will
make UW equivalent to the regular writing scheme. For this experiment,
we use Adam optimizer \cite{kingma2014adam} with initial learning
rate and gradient clipping of $\left\{ 0.001,0.0001\right\} $ and
$\left\{ 1,5,10\right\} $, respectively. The metric used to measure
the performance is the average accuracy across decoding steps. For
each configuration of hyper-parameters, we run the experiment 5 times
and report the mean accuracy with error bars.

Figs. \ref{fig:The-test-accuracy}(a) and (c) depict the performance
of UW and regular writing under different configurations. In any case,
UW boosts the prediction accuracy of MANNs. The performance gain can
be seen clearly when the compression ratio is between $10-40\%$.
This is expected since when the compression ratio is too small or
too big, UW converges to regular writing. Interestingly, increasing
the memory size does not always improve the performance, as in the
case of NTM with RNN controllers. Perhaps, learning to attend to many
memory slots is tricky for some task given limited amount of training
data. This supports the need to apply UW to MANN with moderate memory
size. We also conduct experiments to verify the benefit of using UW
for bigger memory. The results can be found in Appendix \ref{subsec:UW-performance-on}. 

We also measure the speed-up of training time when applying UW on
DNC and NTM, which is illustrated in Figs. \ref{fig:The-test-accuracy}(b)
and (d). The result shows that with UW, the training time can drop
up to $60\%$ for DNC and $28\%$ for NTM, respectively. As DNC is
more complicated than NTM, using UW to reduce memory access demonstrates
clearer speed-up in training (similar behavior can be found for testing
time).

\subsection{Synthetic Memorization}

Here we address a broader range of baselines on two synthetic memorization
tasks, which are the sequence copy and reverse. In these tasks, there
is no discrimination amongst timesteps so the model's goal is to learn
to compress the input efficiently for later retrieval. We experiment
with different sequence lengths of 50 and 100 timesteps. Other details
are the same as the previous double task except that we fix the learning
rate and gradient clipping to 0.001 and 10, respectively. The standard
baselines include LSTM, NTM and DNC. All memory-augmented models have
the same memory size of $4$ slots, corresponding to compression ratio
of $10\%$ and $5\%$, respectively. We aim at this range of compression
ratio to match harsh practical requirements. UW and CUW (cache size
$L=5$) are built upon the DNC, which from our previous observations,
works best for given compression ratios. We choose different dimensions
$N_{h}$ for the hidden vector of the controllers to ensure the model
sizes are approximately equivalent. To further verify that our UW
is actually the optimal writing strategy, we design a new baseline,
which is DNC with random irregular writing strategy (RW). The write
is sampled from a binomial distribution with $p=\left(D+1\right)/T$
(equivalent to compression ratio). After sampling, we conduct the
training for that policy. The final performances of RW are taken average
from 3 different random policies' results.

The performance of the models is listed in Table \ref{tab:mem_test}.
As clearly seen, UW is the best performer for the pure memorization
tests. This is expected from the theory as all timesteps are importantly
equivalent. Local attention mechanism in CUW does not help much in
this scenario and thus CUW finishes the task as the runner-up. Reverse
seems to be easier than copy as the models tend to ``remember''
more the last-seen timesteps whose contributions $\lambda^{T-t}$
remains significant. In both cases, other baselines including random
irregular and regular writing underperform our proposed models by
a huge margin. 

\begin{table}
\begin{centering}
\begin{tabular}{|c|c|c|c|c||c|c|}
\hline 
\multirow{2}{*}{Model} & \multirow{2}{*}{$N_{h}$} & \multirow{2}{*}{\# parameter} & \multicolumn{2}{c||}{Copy} & \multicolumn{2}{c|}{Reverse}\tabularnewline
\cline{4-7} 
 &  &  & L=50 & L=100 & L=50 & L=100\tabularnewline
\hline 
\hline 
LSTM & 125 & 103,840 & 15.6 & 12.7 & 49.6 & 26,1\tabularnewline
\hline 
NTM & 100 & 99,112 & 40.1 & 11.8 & 61.1 & 20.3\tabularnewline
\hline 
DNC & 100 & 98,840 & 68.0 & 44.2 & 65.0 & 54.1\tabularnewline
\hline 
DNC+RW & 100 & 98,840 & 47.6 & 37.0 & 70.8 & 50.1\tabularnewline
\hline 
\hline 
DNC+UW & 100 & 98,840 & \textbf{97.7} & \textbf{69.3} & \textbf{100} & \textbf{79.5}\tabularnewline
\hline 
DNC+CUW & 95  & 96,120 & 83.8 & 55.7 & 93.3 & 55.4\tabularnewline
\hline 
\end{tabular}
\par\end{centering}
\caption{Test accuracy (\%) on synthetic memorization tasks. MANNs have 4 memory
slots.\label{tab:mem_test}}
\end{table}

\subsection{Synthetic Reasoning}

Tasks in the real world rarely involve just memorization. Rather,
they require the ability to selectively remember the input data and
synthesize intermediate computations. To investigate whether our proposed
writing schemes help the memory-augmented models handle these challenges,
we conduct synthetic reasoning experiments which include add and max
tasks. In these tasks, each number in the output sequence is the sum
or the maximum of two numbers in the input sequence. The pairing is
fixed as: $y_{t}=\frac{x_{t}+x_{T-t}}{2},t=\overline{1,\left\lfloor \frac{T}{2}\right\rfloor }$
for add task and $y_{t}=\max\left(x_{2t},x_{2t+1}\right),t=\overline{1,\left\lfloor \frac{T}{2}\right\rfloor }$
for max task, respectively. The length of the output sequence is thus
half of the input sequence. A brief overview of input/output format
for these tasks can be found in Appendix \ref{subsec:Summary-of-synthetic}.
We deliberately use local (max) and distant (add) pairing rules to
test the model under different reasoning strategies. The same experimental
setting as in the previous section is applied except for the data
sample range for the max task, which is $\left[1,50\right]$\footnote{With small range like $\left[1,10\right]$, there is no much difference
in performance amongst models}. LSTM and NTM are excluded from the baselines as they fail on these
tasks. 

Table \ref{tab:reason_task} shows the testing results for the reasoning
tasks. Since the memory size is small compared to the number of events,
regular writing or random irregular writing cannot compete with the
uniform-based writing policies. Amongst all baselines, CUW demonstrates
superior performance in both tasks thanks to its local attention mechanism.
It should be noted that the timesteps should not be treated equally
in these reasoning tasks. The model should weight a timestep differently
based on either its content (max task) or location (add task) and
maintain its memory for a long time by following uniform criteria.
CUW is designed to balance the two approaches and thus it achieves
better performance. Further insights into memory operations of these
models are given in Appendix \ref{subsec:Memory-writing-behaviours}. 

\begin{table}
\begin{centering}
\begin{tabular}{|c|c|c||c|c|}
\hline 
\multirow{2}{*}{Model} & \multicolumn{2}{c||}{Add} & \multicolumn{2}{c|}{Max}\tabularnewline
\cline{2-5} 
 & L=50 & L=100 & L=50 & L=100\tabularnewline
\hline 
\hline 
DNC & 83.8 & 22.3 & 59.5 & 27.4\tabularnewline
\hline 
DNC+RW & 83.0 & 22.7 & 59.7 & 36.5\tabularnewline
\hline 
\hline 
DNC+UW & 84.8 & 50.9 & 71.7 & 66.2\tabularnewline
\hline 
DNC+CUW & \textbf{94.4} & \textbf{60.1} & \textbf{82.3} & \textbf{70.7}\tabularnewline
\hline 
\end{tabular}
\par\end{centering}
\caption{Test accuracy (\%) on synthetic reasoning tasks. MANNs have 4 memory
slots.\label{tab:reason_task}}
\end{table}

\subsection{Synthetic Sinusoidal Regression }

In real-world settings, sometimes a long sequence can be captured
and fully reconstructed by memorizing some of its feature points.
For examples, a periodic function such as sinusoid can be well-captured
if we remember the peaks of the signal. By observing the peaks, we
can deduce the frequency, amplitude, phase and thus fully reconstructing
the function. To demonstrate that UW and CUW are useful for such scenarios,
we design a sequential continuation task, in which the input is a
sequence of sampling points across some sinusoid: $y=5+A\sin(2\pi fx+\varphi)$.
Here, $A\sim\mathcal{U}\left(1,5\right)$, $f\sim\mathcal{U}\left(10,30\right)$
and $\varphi\sim\mathcal{U}\left(0,100\right)$. After reading the
input $y=\left\{ y_{t}\right\} _{t=1}^{T}$, the model have to generate
a sequence of the following points in the sinusoid. To ensure the
sequence $y$ varies and covers at least one period of the sinusoid,
we set $x=\left\{ x_{t}\right\} _{t=1}^{T}$ where $x_{i}=\left(t+\epsilon_{1}\right)/1000$,
$\epsilon_{1}\sim\mathcal{U}\left(-1,1\right)$. The sequence length
for both input and output is fixed to $T=100$. The experimental models
are LSTM, DNC, UW and CUW (built upon DNC). For each model, optimal
hyperparameters including learning rate and clipping size are tuned
with 10,000 generated sinusoids. The memories have $4$ slots and
all baselines have similar parameter size. We also conduct the experiment
with noisy inputs by adding a noise $\epsilon_{2}\sim\mathcal{U}\left(-2,2\right)$
to the input sequence $y$. This increases the difficulty of the task.
The loss is the average of mean square error (MSE) over decoding timesteps.

\begin{figure}
\begin{centering}
\includegraphics[width=1\textwidth]{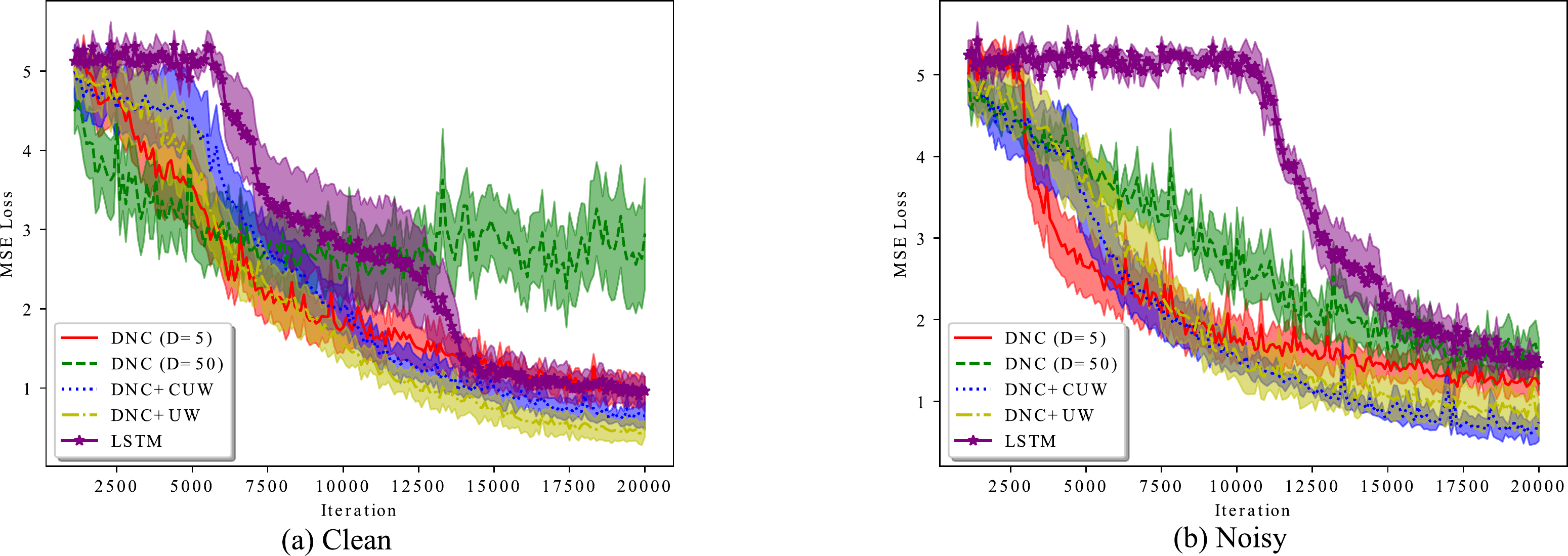}
\par\end{centering}
\caption{Learning curves of models in clean (a) and noisy (b) sinusoid regression
experiment.\label{fig:Training-curves-of}}

\end{figure}

We plot the mean learning curves with error bars over 5 runnings for
sinusoidal regression task under clean and noisy condition in Figs.
\ref{fig:Training-curves-of}(a) and (b), respectively. Regular writing
DNC learns fast at the beginning, yet soon saturates and approaches
the performance of LSTM ($MSE=1.05$ and $1.39$ in clean and noisy
condition, respectively). DNC performance does not improve much as
we increase the memory size to $50$, which implies the difficulty
in learning with big memory. Although UW starts slower, it ends up
with lower errors than DNC and perform slightly better than CUW in
clean condition ($MSE=0.44$ for UW and $0.61$ for CUW). CUW demonstrates
competitive performance against other baselines, approaching to better
solution than UW for noisy task where the model should discriminate
the timesteps ($MSE=0.98$ for UW and $0.55$ for CUW). More visualizations
can be found in Appendix \ref{subsec:Visualizations-of-model}. 

\subsection{Flatten Image Recognition}

We want to compare our proposed models with DNC and other methods
designed to help recurrent networks learn longer sequence. The chosen
benchmark is a pixel-by-pixel image classification task on MNIST in
which pixels of each image are fed into a recurrent model sequentially
before a prediction is made. In this task, the sequence length is
fixed to 768 with highly redundant timesteps (black pixels). The training,
validation and testing sizes are 50,000, 10,000 and 10,000, respectively.
We test our models on both versions of non-permutation (MNIST) and
permutation (pMNIST). More details on the task and data can be found
in \cite{le2015simple}. $\joinrel$For DNC, we try with several memory
slots from $\left\{ 15,30,60\right\} $ and report the best results.
For UW and CUW, memory size is fixed to 15 and cache size $L$ is
set to 10. The controllers are implemented as single layer GRU with
$100$-dimensional hidden vector. To optimize the models, we use RMSprop
with initial learning rate of 0.0001. 

Table \ref{tab:mnist} shows that DNC underperforms r-LSTM, which
indicates that regular DNC with big memory finds it hard to beat LSTM-based
methods. After applying UW, the results get better and with CUW, it
shows significant improvement over r-LSTM and demonstrates competitive
performance against dilated-RNNs models. Notably, dilated-RNNs use
9 layers in their experiments compared to our singer layer controller.
Furthermore, our models exhibit more consistent performance than dilated-RNNs.
For completeness, we include comparisons between CUW and non-recurrent
methods in Appendix \ref{subsec:Comparsion-with-non-recurrent}

\begin{table}
\begin{centering}
\begin{tabular}{|l|c|c|}
\hline 
Model & MNIST & pMNIST\tabularnewline
\hline 
\hline 
iRNN$^{\dagger}$  & 97.0 & 82.0\tabularnewline
\hline 
uRNN$^{\circ}$ & 95.1 & 91.4\tabularnewline
\hline 
r-LSTM Full BP$^{\star}$ & 98.4 & 95.2\tabularnewline
\hline 
Dilated-RNN$^{\blacklozenge}$ & 95.5 & 96.1\tabularnewline
\hline 
Dilated-GRU$^{\blacklozenge}$ & \textbf{99.2} & 94.6\tabularnewline
\hline 
DNC & 98.1 & 94.0\tabularnewline
\hline 
\hline 
DNC+UW & 98.6 & 95.6\tabularnewline
\hline 
DNC+CUW & 99.1 & \textbf{96.3}\tabularnewline
\hline 
\end{tabular}
\par\end{centering}
\caption{Test accuracy (\%) on MNIST, pMNIST. Previously reported results are
from \cite{le2015simple}$^{\dagger}$, \cite{arjovsky2016unitary}$^{\circ}$,
\cite{trinh2018learning}$^{\star}$, and \cite{chang2017dilated}$^{\blacklozenge}$.
\label{tab:mnist}}
\end{table}

\subsection{Document Classification}

To verify our proposed models in real-world applications, we conduct
experiments on document classification task. In the task, the input
is a sequence of words and the output is the classification label.
Following common practices in \cite{yogatama2017generative,seo2018neural},
each word in the document is embedded into a $300$-dimensional vector
using Glove embedding \cite{pennington2014glove}. We use RMSprop
for optimization, with initial learning rate of 0.0001. Early-stop
training is applied if there is no improvement after 5 epochs in the
validation set. Our UW and CUW are built upon DNC with single layer
$512$-dimensional LSTM controller and the memory size is chosen in
accordance with the average length of the document, which ensures
$10-20\%$ compression ratio. The cache size for CUW is fixed to 10.
The datasets used in this experiment are common big datasets where
the number of documents is between 120,000 and 1,400,000 with maximum
of 4,392 words per document (see Appendix \ref{subsec:Details-on-document}
for further details). The baselines are recent state-of-the-arts in
the domain, some of which are based on recurrent networks such as
D-LSTM \cite{yogatama2017generative} and Skim-LSTM \cite{seo2018neural}.
We exclude DNC from the baselines as it is inefficient to train the
model with big document datasets. 

Our results are reported in Table \ref{tab:text}. On five datasets
out of six, our models beat or match the best published results. For
IMDb dataset, our methods outperform the best recurrent model (Skim-LSTM).
The performance gain is competitive against that of the state-of-the-arts.
In most cases, CUW is better than UW, which emphasizes the importance
of relaxing the timestep equality assumption in practical situations.
Details results across different runs for our methods are listed in
Appendix \ref{subsec:Document-classification-detailed}. 

\begin{table}
\begin{centering}
\begin{tabular}{|l|c|c|c|c|c|c|}
\hline 
Model & AG & IMDb\tablefootnote{Methods that use semi-supervised training to achieve higher accuracy
are not listed.} & Yelp P. & Yelp F. & DBP & Yah. A.\tabularnewline
\hline 
\hline 
VDCNN$^{\bullet}$  & 91.3 & - & 95.7 & 64.7 & 98.7 & 73.4\tabularnewline
\hline 
D-LSTM$^{\ast}$ & - & - & 92.6 & 59.6 & 98.7 & \textit{73.7}\tabularnewline
\hline 
Standard LSTM$^{\ddagger}$ & 93.5 & 91.1 & - & - & - & -\tabularnewline
\hline 
Skim-LSTM$^{\ddagger}$ & \textit{93.6} & 91.2 & - & - & - & -\tabularnewline
\hline 
Region Embedding$^{\blacktriangle}$ & 92.8 & - & \textbf{\textit{96.4}} & \textit{64.9} & \textit{98.9} & \textit{73.7}\tabularnewline
\hline 
\hline 
DNC+UW & 93.7 & \textbf{91.4} & \textbf{96.4} & 65.3 & \textbf{99.0} & 74.2\tabularnewline
\hline 
DNC+CUW & \textbf{93.9} & 91.3 & \textbf{96.4} & \textbf{65.6} & \textbf{99.0} & \textbf{74.3}\tabularnewline
\hline 
\end{tabular}
\par\end{centering}
\caption{Document classification accuracy (\%) on several datasets. Previously
reported results are from \cite{conneau2016very}$^{\bullet}$, \cite{yogatama2017generative}$^{\ast}$,
\cite{seo2018neural}$^{\ddagger}$ and \cite{qiao2018a}$^{\blacktriangle}$.
We use italics to denote the best published and bold the best records.\label{tab:text}}

\end{table}

%% file: related.tex
Traditional recurrent models such as RNN/LSTM \cite{elman1990finding,hochreiter1997long}
exhibit some weakness that prevent them from learning really long
sequences. The reason is mainly due to the vanishing gradient problem
\cite{pascanu2013difficulty} or to be more specific, the exponential
decay of input value over time. One way to overcome this problem is
enforcing the exponential decay factor close to one by putting a unitary
constraint on the recurrent weight \cite{arjovsky2016unitary,wisdom2016full}.
Although this approach is theoretically motivated, it restricts the
space of learnt parameters. 

More relevant to our work, the idea of using less or adaptive computation
for good has been proposed in \cite{graves2016adaptive,yu2017learning,yu2018fast,seo2018neural}.
Most of these works are based on the assumption that some of timesteps
in a sequence are unimportant and thus can be ignored to reduce the
cost of computation and increase the performance of recurrent networks.
Different form our approach, these methods lack theoretical supports
and do not directly aim to solve the problem of memorizing long-term
dependencies. 

Dilated RNN \cite{chang2017dilated} is another RNN-based proposal
which improves long-term learning by stacking multiple dilated recurrent
layers with hierarchical skip-connections. This theoretically guarantees
the mean recurrent length and shares with our method the idea to construct
a measurement on memorization capacity of the system and propose solutions
to optimize it. The difference is that our system is memory-augmented
neural networks while theirs is multi-layer RNNs, which leads to totally
different optimization problems. 

Recent researches recommend to replace traditional recurrent models
by other neural architectures to overcome the vanishing gradient problem.
The Transformer \cite{vaswani2017attention} attends to all timesteps
at once, which ensures instant access to distant timestep yet requires
quadratic computation and physical memory proportional to the sequence
length. Memory-augmented neural networks (MANNs), on the other hand,
learn to establish a limited-size memory and attend to the memory
only, which is scalable to any-length sequence. Compared to others,
MANNs resemble both computer architecture design and human working
memory \cite{logie2014visuo}. However, the current understanding
of the underlying mechanisms and theoretical foundations for MANN
are still limited.

Recent works on MANN rely almost on reasonable intuitions. Some introduce
new addressing mechanisms such as location-based \cite{graves2014neural},
least-used \cite{santoro2016meta} and order-based \cite{graves2016hybrid}.
Others focus on the scalability of MANN by using sparse memory access
to avoid attending to a large number of memory slots \cite{rae2016scaling}.
These problems are different from ours which involves MANN memorization
capacity optimization. 

Our local optimal solution to this problem is related to some known
neural caching \cite{grave2017improving,grave2017unbounded,yogatama2018memory}
in terms of storing recent hidden states for later encoding uses.
These methods either aim to create structural bias to ease the learning
process \cite{yogatama2018memory} or support large scale retrieval
\cite{grave2017unbounded}. These are different from our caching purpose,
which encourages overwriting and relaxes the equal contribution assumption
of the optimal solution. Also, the details of implementation are different
as ours uses local memory-augmented attention mechanisms.

%% file: discuss.tex
We have introduced Uniform Writing (UW) and Cached Uniform Writing
(CUW) as faster solutions for longer-term memorization in MANNs. With
a comprehensive suite of synthetic and practical experiments, we provide
strong evidences that our simple writing mechanisms are crucial to
MANNs to reduce computation complexity and achieve competitive performance
in sequence modeling tasks. In complement to the experimental results,
we have proposed a meaningful measurement on MANN memory capacity
and provided theoretical analysis showing the optimality of our methods.
Further investigations to tighten the measurement bound will be the
focus of our future work.

%% file: appendix.tex
\subsection{Derivation on the bound inequality in linear dynamic system\label{subsec:Derivation-on-theRNN}}

The linear dynamic system hidden state is described by the following
recursive equation:

\[
h_{t}=Wx_{t}+Uh_{t-1}+b
\]
By induction,

\[
h_{t}=\stackrel[i=1]{t}{\sum}U^{t-i}Wx_{i}+C
\]
where $C$ is some constant with respect to $x_{i}$. In this case,
$\frac{\partial h_{t}}{\partial x_{i}}=U^{t-i}W$. By applying norm
sub-multiplicativity\footnote{If not explicitly stated otherwise, norm refers to any consistent
matrix norm which satisfies sub-multiplicativity. }, 

\begin{align*}
c_{i-1,t} & =\left\Vert U^{t-i+1}W\right\Vert \\
 & \leq\left\Vert U\right\Vert \left\Vert U^{t-i}W\right\Vert \\
 & =\left\Vert U\right\Vert c_{i,t}
\end{align*}
That is, $\lambda_{c}=\left\Vert U\right\Vert $.

\subsection{Derivation on the bound inequality in standard RNN\label{subsec:Derivation-on-theRNN-1}}

The standard RNN hidden state is described by the following recursive
equation:

\[
h_{t}=\tanh\left(Wx_{t}+Uh_{t-1}+b\right)
\]
From $\frac{\partial h_{t}}{\partial x_{i}}=\frac{\partial h_{t}}{\partial h_{t-1}}\frac{\partial h_{t-1}}{\partial x_{i}}$,
by induction, 

\[
\frac{\partial h_{t}}{\partial x_{i}}=\left(\stackrel[j=i+1]{t}{\prod}\frac{\partial h_{j}}{\partial h_{j-1}}\right)\frac{\partial h_{i}}{\partial x_{i}}=\left(\stackrel[j=i]{t}{\prod}diag\left(\tanh^{\prime}\left(a_{j}\right)\right)\right)U^{t-i}W
\]
where $a_{j}=Wx_{j}+Uh_{j-1}+b$ and $diag\left(\cdot\right)$ converts
a vector into a diagonal matrix. As $0\leq\tanh^{\prime}\left(x\right)=1-\tanh\left(x\right)^{2}\leq1$,
$\left\Vert diag\left(\tanh^{\prime}\left(X\right)\right)\right\Vert $
is bounded by some value $B$. By applying norm sub-multiplicativity, 

\begin{align*}
c_{i-1,t} & =\left\Vert \left[\stackrel[j=i]{t}{\prod}diag\left(\tanh^{\prime}\left(a_{j}\right)\right)\right]diag\left(\tanh^{\prime}\left(a_{i-1}\right)\right)U^{t-i+1}W\right\Vert \\
 & \leq\left\Vert U\right\Vert \left\Vert \stackrel[j=i]{t}{\prod}diag\left(\tanh^{\prime}\left(a_{j}\right)\right)U^{t-i}W\right\Vert \left\Vert diag\left(\tanh^{\prime}\left(a_{i-1}\right)\right)\right\Vert \\
 & =B\left\Vert U\right\Vert c_{i,t}
\end{align*}
That is, $\lambda_{c}=B\left\Vert U\right\Vert $.

\subsection{Derivation on the bound inequality in LSTM\label{subsec:Derivation-on-theLSTM}}

For the case of LSTM, the recursive equation reads:
\begin{align*}
c_{t} & =\sigma\left(U_{f}x_{t}+W_{f}h_{t-1}+b_{f}\right)\odot c_{t-1}\\
 & +\sigma\left(U_{i}x_{t}+W_{i}h_{t-1}+b_{i}\right)\odot\tanh\left(U_{z}x_{t}+W_{z}h_{t-1}+b_{z}\right)\\
h_{t} & =\sigma\left(U_{o}x_{t}+W_{o}h_{t-1}+b_{o}\right)\odot\tanh\left(c_{t}\right)
\end{align*}

Taking derivatives,

\begin{align*}
\frac{\partial h_{j}}{\partial h_{j-1}} & =\sigma^{\prime}\left(o_{j}\right)\tanh\left(c_{j}\right)W_{o}+\sigma\left(o_{j}\right)\tanh^{\prime}\left(c_{j}\right)\sigma^{\prime}\left(f_{j}\right)c_{j-1}W_{f}\\
 & +\sigma\left(o_{j}\right)\tanh^{\prime}\left(c_{j}\right)\sigma^{\prime}\left(i_{j}\right)\tanh\left(z_{j}\right)W_{i}+\sigma\left(o_{j}\right)\tanh^{\prime}\left(c_{j}\right)\sigma\left(i_{j}\right)\tanh^{\prime}\left(z_{j}\right)W_{z}\\
\frac{\partial h_{j-1}}{\partial x_{j-1}} & =\sigma^{\prime}\left(o_{j-1}\right)\tanh\left(c_{j-1}\right)U_{o}+\sigma\left(o_{j-1}\right)\tanh^{\prime}\left(c_{j-1}\right)\sigma^{\prime}\left(f_{j-1}\right)c_{j-2}U_{f}\\
 & +\sigma\left(o_{j-1}\right)\tanh^{\prime}\left(c_{j-1}\right)\sigma^{\prime}\left(i_{j-1}\right)\tanh\left(z_{j-1}\right)U_{i}+\sigma\left(o_{j-1}\right)\tanh^{\prime}\left(c_{j-1}\right)\sigma\left(i_{j-1}\right)\tanh^{\prime}\left(z_{j-1}\right)U_{z}\\
\frac{\partial h_{j}}{\partial x_{j}} & =\sigma^{\prime}\left(o_{j}\right)\tanh\left(c_{j}\right)U_{o}+\sigma\left(o_{j}\right)\tanh^{\prime}\left(c_{j}\right)\sigma^{\prime}\left(f_{j}\right)c_{j-1}U_{f}\\
 & +\sigma\left(o_{j}\right)\tanh^{\prime}\left(c_{j}\right)\sigma^{\prime}\left(i_{j}\right)\tanh\left(z_{j}\right)U_{i}+\sigma\left(o_{j}\right)\tanh^{\prime}\left(c_{j}\right)\sigma\left(i_{j}\right)\tanh^{\prime}\left(z_{j}\right)U_{z}
\end{align*}
where $o_{j}$ denotes the value in the output gate at $j$-th timestep
(similar notations are used for input gate ($i_{j}$), forget gate
($f_{j}$) and cell value ($z_{j}$)) and ``non-matrix'' terms actually
represent diagonal matrices corresponding to these terms. Under the
assumption that $h_{0}$=0, we then make use of the results in \cite{miller2018recurrent}
stating that $\left\Vert c_{t}\right\Vert _{\infty}$ is bounded for
all $t$. By applying $l_{\infty}$-norm sub-multiplicativity and
triangle inequality, we can show that

\[
\frac{\partial h_{j}}{\partial h_{j-1}}\frac{\partial h_{j-1}}{\partial x_{j-1}}=M\frac{\partial h_{j}}{\partial x_{j}}+N
\]

with

\begin{align*}
\left\Vert M\right\Vert _{\infty} & \leq1/4\left\Vert W_{o}\right\Vert _{\infty}+1/4\left\Vert W_{f}\right\Vert _{\infty}\left\Vert c_{j}\right\Vert _{\infty}+1/4\left\Vert W_{i}\right\Vert _{\infty}+\left\Vert W_{z}\right\Vert _{\infty}=B_{m}\\
\left\Vert N\right\Vert _{\infty} & \leq1/16\left\Vert W_{o}U_{i}\right\Vert _{\infty}+1/16\left\Vert W_{o}U_{f}\right\Vert _{\infty}\left(\left\Vert c_{j}\right\Vert _{\infty}+\left\Vert c_{j-1}\right\Vert _{\infty}\right)+1/4\left\Vert W_{o}U_{z}\right\Vert _{\infty}\\
 & +1/16\left\Vert W_{i}U_{o}\right\Vert _{\infty}+1/16\left\Vert W_{i}U_{f}\right\Vert _{\infty}\left(\left\Vert c_{j}\right\Vert _{\infty}+\left\Vert c_{j-1}\right\Vert _{\infty}\right)+1/4\left\Vert W_{i}U_{z}\right\Vert _{\infty}\\
 & +1/16\left(\left\Vert W_{f}U_{o}\right\Vert _{\infty}+1/16\left\Vert W_{f}U_{i}\right\Vert _{\infty}+1/4\left\Vert W_{f}U_{z}\right\Vert _{\infty}\right)\left(\left\Vert c_{j}\right\Vert _{\infty}+\left\Vert c_{j-1}\right\Vert _{\infty}\right)\\
 & +1/4\left\Vert W_{z}U_{o}\right\Vert _{\infty}+1/4\left\Vert W_{z}U_{f}\right\Vert _{\infty}\left(\left\Vert c_{j}\right\Vert _{\infty}+\left\Vert c_{j-1}\right\Vert _{\infty}\right)+1/4\left\Vert W_{z}U_{i}\right\Vert _{\infty}\\
 & =B_{n}
\end{align*}

By applying $l_{\infty}$-norm sub-multiplicativity and triangle inequality,

\begin{align*}
c_{i-1,t} & =\left\Vert \stackrel[j=i+1]{t}{\prod}\frac{\partial h_{j}}{\partial h_{j-1}}\frac{\partial h_{i}}{\partial h_{i-1}}\frac{\partial h_{i-1}}{\partial x_{i-1}}\right\Vert _{\infty}\\
 & =\left\Vert \stackrel[j=i+1]{t}{\prod}\frac{\partial h_{j}}{\partial h_{j-1}}\left(\frac{\partial h_{i}}{\partial x_{i}}m+n\right)\right\Vert _{\infty}\\
 & \leq B_{m}c_{i,t}+B_{n}\stackrel[j=i+1]{t}{\prod}\left\Vert \frac{\partial h_{j}}{\partial h_{j-1}}\right\Vert _{\infty}
\end{align*}

As LSTM is $\lambda$-contractive with $\lambda<1$ in the $l_{\infty}$-norm
(readers are recommended to refer to \cite{miller2018recurrent} for
proof), which implies $\left\Vert \frac{\partial h_{j}}{\partial h_{j-1}}\right\Vert _{\infty}<1$,
$B_{n}\stackrel[j=i+1]{t}{\prod}\left\Vert \frac{\partial h_{j}}{\partial h_{j-1}}\right\Vert _{\infty}\rightarrow0$
as $t-i\rightarrow\infty$. For $t-i<\infty$, under the assumption
that $\frac{\partial h_{j}}{\partial x_{j}}\neq0$, we can always
find some value $B<\infty$ such that $c_{i-1,t}\leq Bc_{i,t}$. For
$t-i\rightarrow\infty$, $\lambda_{c}\rightarrow B_{m}$. That is,
$\lambda_{c}=\max\left(B_{m},B\right)$.

\subsection{Proof of theorem \ref{thm:The-average-amount}\label{subsec:Proof-of-theorem-1}}
\begin{proof}
Given that $\lambda_{c}c_{i,t}\geq c_{i-1,t}$ with some $\lambda_{c}\in\mathbb{R^{+}}$,
we can use $c_{t,t}\lambda_{c}^{t-i}$ as the upper bound on $c_{i,t}$
with $i=\overline{1,t}$, respectively. Therefore,

\[
f(0)\leq\stackrel[t=1]{T}{\sum}c_{t,T}\leq c_{T,T}\stackrel[t=1]{T}{\sum}\lambda_{c}^{T-t}=f\left(\lambda_{c}\right)
\]
where $f(\lambda)=c_{T,T}\stackrel[t=1]{T}{\sum}\lambda^{T-t}$ is
continuous on $\mathbb{R^{+}}$. According to intermediate value theorem,
there exists $\lambda\in\left(0,\lambda_{c}\right]$ such that $c_{T,T}\stackrel[t=1]{T}{\sum}\lambda^{T-t}=\stackrel[t=1]{T}{\sum}c_{t,T}$.
\end{proof}

\subsection{Proof of theorem \ref{thm:Under-the-assumption}\label{subsec:Proof-of-theorem-2}}
\begin{proof}
According to Theorem \ref{thm:The-average-amount}, there exists some
$\lambda_{i}\in\mathbb{R^{+}}$such that the summation of contribution
stored between $K_{i}$ and $K_{i+1}$ can be quantified as $c_{K_{i+1},K_{i+1}}\stackrel[t=K_{i}]{K_{i+1}}{\sum}\lambda_{i}^{K_{i+1}-t}$
(after ignoring contributions before $K_{i}$-th timestep for simplicity).
Let denote $P(\lambda)=\stackrel[t=K_{i}]{K_{i+1}}{\sum}\lambda^{K_{i+1}-t}$,
we have $P^{\prime}\left(\lambda\right)>0,\mathbb{\forall\lambda\in R^{+}}$.
Therefore, $P(\lambda_{i})\geq P\left(\underset{i}{\min}\left(\lambda_{i}\right)\right)$.
Let $C=\underset{i}{\min}\left(c_{i,i}\right)$ and $\lambda=\underset{i}{\min}\left(\lambda_{i}\right)$,
the average contribution stored in a MANN has a lower bound quantified
as $I_{\lambda}$, where $I_{\lambda}=C\frac{\stackrel[t=1]{K_{1}}{\sum}\lambda^{K_{1}-t}+\stackrel[t=K_{1}+1]{K_{2}}{\sum}\lambda^{K_{2}-t}+...+\stackrel[t=K_{D-1}+1]{K_{D}}{\sum}\lambda^{K_{D}-t}+\stackrel[t=K_{D}+1]{T}{\sum}\lambda^{T-t}}{T}$.
\end{proof}

\subsection{Proof of theorem \ref{thm:Given-the-number}\label{subsec:Proof-of-theorem}}
\begin{proof}
The second-order derivative of $f_{\lambda}\left(x\right)$ reads:

\begin{equation}
f_{\lambda}^{\prime\prime}\left(x\right)=-\frac{\left(\ln\lambda\right)^{2}}{1-\lambda}\lambda^{x}
\end{equation}

We have $f_{\lambda,}^{\prime\prime}\left(x\right)\leq0$ with $\forall x\in\mathbb{R^{+}}$
and $1>\lambda>0$, so $f_{\lambda}\left(x\right)$ is a concave function.
Thus, we can apply Jensen inequality as follows: 
\begin{equation}
\frac{1}{D+1}\stackrel[i=1]{D+1}{\sum}f_{\lambda}(l_{i})\leq f_{\lambda}\left(\stackrel[i=1]{D+1}{\sum}\frac{1}{D+1}l_{i}\right)=f_{\lambda}\left(\frac{T}{D+1}\right)\label{eq:jensen}
\end{equation}
Equality holds if and only if $l_{1}=l_{2}=...=l_{D+1}=\frac{T}{D+1}$.
We refer to this as\textit{ Uniform Writing} strategy. By plugging
the optimal values of $l_{i}$, we can derive the maximized average
contribution as  follows:

\begin{equation}
I_{\lambda}max\equiv g_{\lambda}\left(T,D\right)=\frac{C\left(D+1\right)}{T}\left(\frac{1-\lambda^{\frac{T}{D+1}}}{1-\lambda}\right)
\end{equation}
When $\lambda=1$, $I_{\lambda}=\frac{C}{T}\stackrel[i=1]{D+1}{\sum}l_{i}=C$.
This is true for all writing strategies. Thus, Uniform Writing is
optimal for $0<\lambda\leq1$.
\end{proof}
We can show that this solution is also optimal for the case $\lambda>1$.
As $f_{\lambda}^{\prime\prime}\left(x\right)>0$ with $\forall x\in\mathbb{R^{+}};\lambda>1$,
$f_{\lambda}\left(x\right)$ is a convex function and Eq. (\ref{eq:jensen})
flips the inequality sign. Thus, $I_{\lambda}$ reaches its minimum
with Uniform Writing. For $\lambda>1$, minimizing $I_{\lambda}$
is desirable to prevent the system from diverging. 

We can derive some properties of function $g$. Let $x=\frac{D+1}{L},g_{\lambda}\left(L,D\right)=g_{\lambda}\left(x\right)=Cx(\frac{\lambda^{\frac{1}{x}}-1}{\lambda-1})$.
We have $g_{\lambda}^{\prime}\left(x\right)=C\lambda\left(1-\lambda^{\frac{1}{x}}\right)\left(x-\ln\lambda\right)>0$
with $0<\lambda\leq1,\forall x\geq0$, so $g_{\lambda}\left(T,D\right)$
is an increasing function if we fix $T$ and let $D$ vary. That explains
why having more memory slots helps improve memorization capacity.
If $D=0$, $g_{\lambda}\left(T,0\right)$ becomes E.q (\ref{eq:d0}).
In this case, MANNs memorization capacity converges to that of recurrent
networks.

\subsection{Summary of synthetic discrete task format\label{subsec:Summary-of-synthetic}}

\begin{table}[H]
\begin{centering}
\begin{tabular}{|c|c|c|}
\hline 
Task & Input & Output\tabularnewline
\hline 
\hline 
Double & $x_{1}x_{2}...x_{T}$ & $x_{1}x_{2}...x_{T}x_{1}x_{2}...x_{T}$\tabularnewline
\hline 
Copy & $x_{1}x_{2}...x_{T}$ & $x_{1}x_{2}...x_{T}$\tabularnewline
\hline 
Reverse & $x_{1}x_{2}...x_{T}$ & $x_{T}x_{T-1}...x_{1}$\tabularnewline
\hline 
Add & $x_{1}x_{2}...x_{T}$ & $\frac{x_{1}+x_{T-1}}{2}\frac{x_{2}+x_{T-2}}{2}...\frac{x_{\left\lfloor T/2\right\rfloor }+x_{\left\lceil T/2\right\rceil }}{2}$\tabularnewline
\hline 
Max & $x_{1}x_{2}...x_{T}$ & $\max\left(x_{1},x_{2}\right)\max\left(x_{3},x_{4}\right)...\max\left(x_{T-1},x_{T}\right)$\tabularnewline
\hline 
\end{tabular}
\par\end{centering}
\caption{Synthetic discrete task's input-output formats. $T$ is the sequence
length.}
\end{table}

\subsection{UW performance on bigger memory\label{subsec:UW-performance-on}}

\begin{table}[H]
\begin{centering}
\begin{tabular}{|c|c|c|}
\hline 
Model & $N_{h}$ & Copy (L=500)\tabularnewline
\hline 
\hline 
DNC & 128 & 24.19\%\tabularnewline
\hline 
DNC+UW & 128 & 81.45\%\tabularnewline
\hline 
\end{tabular}
\par\end{centering}
\caption{Test accuracy (\%) on synthetic copy task. MANNs have 50 memory slots.
Both models are trained with 100,000 mini-batches of size 32.}

\end{table}

\subsection{Memory operating behaviors on synthetic tasks\label{subsec:Memory-writing-behaviours}}

\begin{figure}
\begin{centering}
\includegraphics[width=1\textwidth]{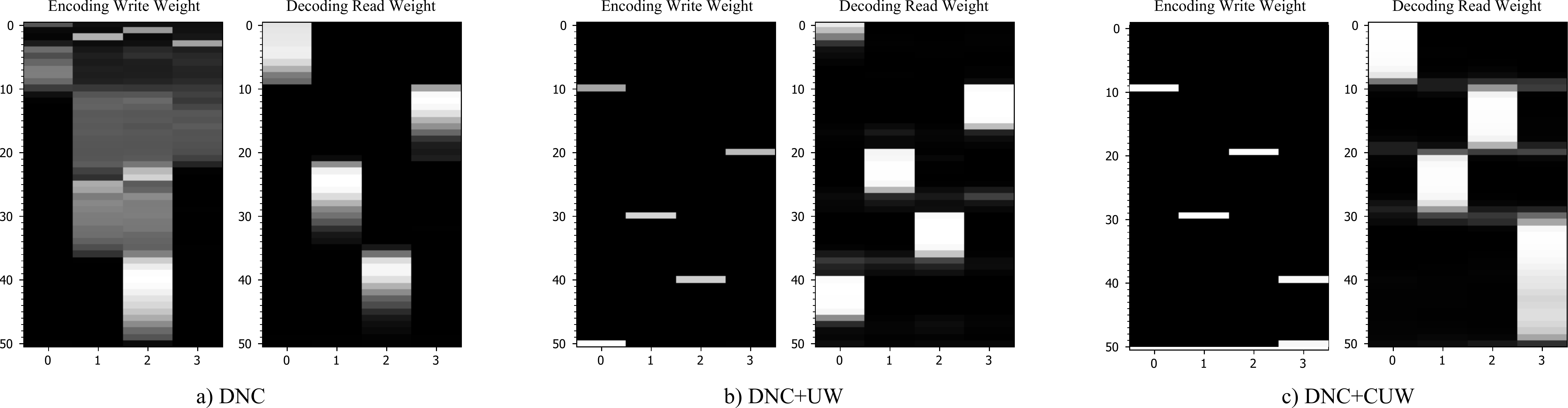}
\par\end{centering}
\caption{Memory operations on copy task in DNC (a), DNC+UW (b) and DNC+CUW(c).
Each row is a timestep and each column is a memory slot.\label{fig:Memory-operations-on}}

\end{figure}

In this section, we pick three models (DNC, DNC+UW and DNC+CUW) to
analyze their memory operating behaviors. Fig. \ref{fig:Memory-operations-on}
visualizes the values of the write weights and read weights for the
copy task during encoding input and decoding output sequence, respectively.
In the copy task, as the sequence length is 50 while the memory size
is 4, one memory slot should contain the accumulation of multiple
timesteps. This principle is reflected in the decoding process in
three models, in which one memory slot is read repeatedly across several
timesteps. Notably, the number of timesteps consecutively spent for
one slot is close to $10$-the optimal interval, even for DNC ( Fig.
\ref{fig:Memory-operations-on}(a)), which implies that the ultimate
rule would be the uniform rule. As UW and CUW are equipped with uniform
writing, their writing patterns follow the rule perfectly. Interestingly,
UW chooses the first written location for the final write (corresponding
to the \textless eos\textgreater{} token) while CUW picks the last
written location. As indicated in Figs. \ref{fig:Memory-operations-on}(b)
and (c), both of them can learn the corresponding reading pattern
for decoding process, which leads to good performances. On the other
hand, regular DNC fails to learn a perfect writing strategy. Except
for the timesteps at the end of the sequence, the timesteps are distributed
to several memory slots while the reading phase attends to one memory
slot repeatedly. This explains why regular DNC cannot compete with
the other two proposed methods in this task. 

\begin{figure}
\begin{centering}
\includegraphics[width=1\textwidth]{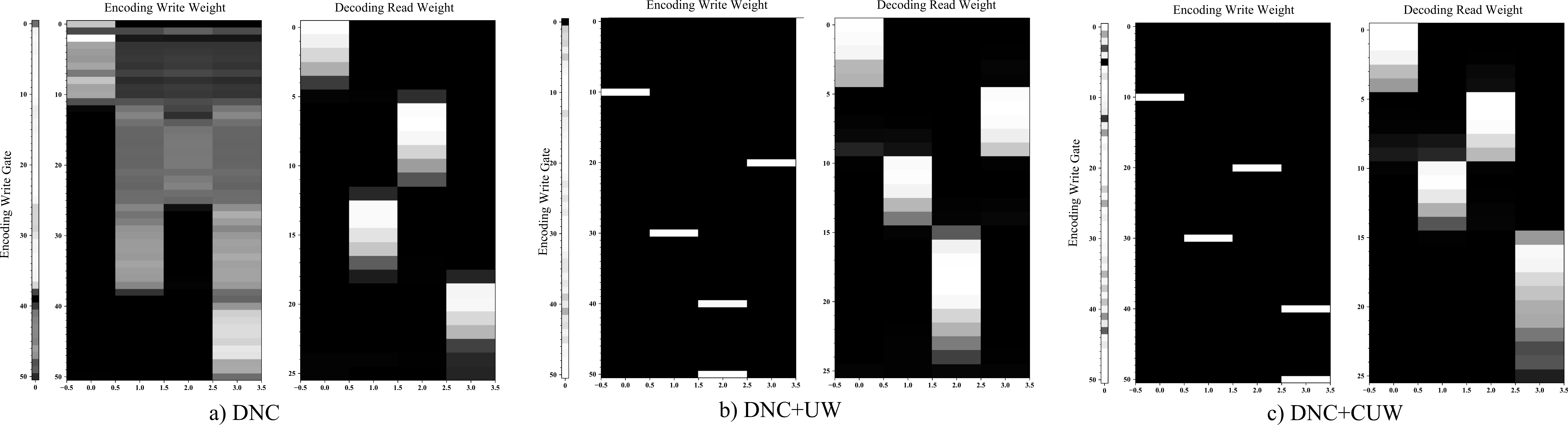}
\par\end{centering}
\caption{Memory operations on max task in DNC (a), DNC+UW (b) and DNC+CUW(c).
Each row is a timestep and each column is a memory slot.\label{fig:Memory-operations-on-1}}
\end{figure}

For the max task, Fig. \ref{fig:Memory-operations-on-1} displays
similar visualization with an addition of write gate during encoding
phase. The write gate indicates how much the model should write the
input at some timestep to the memory. A zero write gate means there
is no writing. For this task, a good model should discriminate between
timesteps and prefer writing the greater ones. As clearly seen in
Fig. \ref{fig:Memory-operations-on-1}(a), DNC suffers the same problem
as in copy task, unable to synchronize encoding writing with decoding
reading. Also, DNC's write gate pattern does not show reasonable discrimination.
For UW (Fig. \ref{fig:Memory-operations-on-1}(b)), it tends to write
every timestep and relies on uniform writing principle to achieve
write/read accordance and thus better results than DNC. Amongst all,
CUW is able to ignore irrelevant timesteps and follows uniform writing
at the same time (see Fig. \ref{fig:Memory-operations-on-1}(c)).


\subsection{Visualizations of model performance on sinusoidal regression tasks\label{subsec:Visualizations-of-model}}

\begin{figure}
\begin{centering}
\noindent\begin{minipage}[t]{1\columnwidth}%
\includegraphics[width=0.3\linewidth]{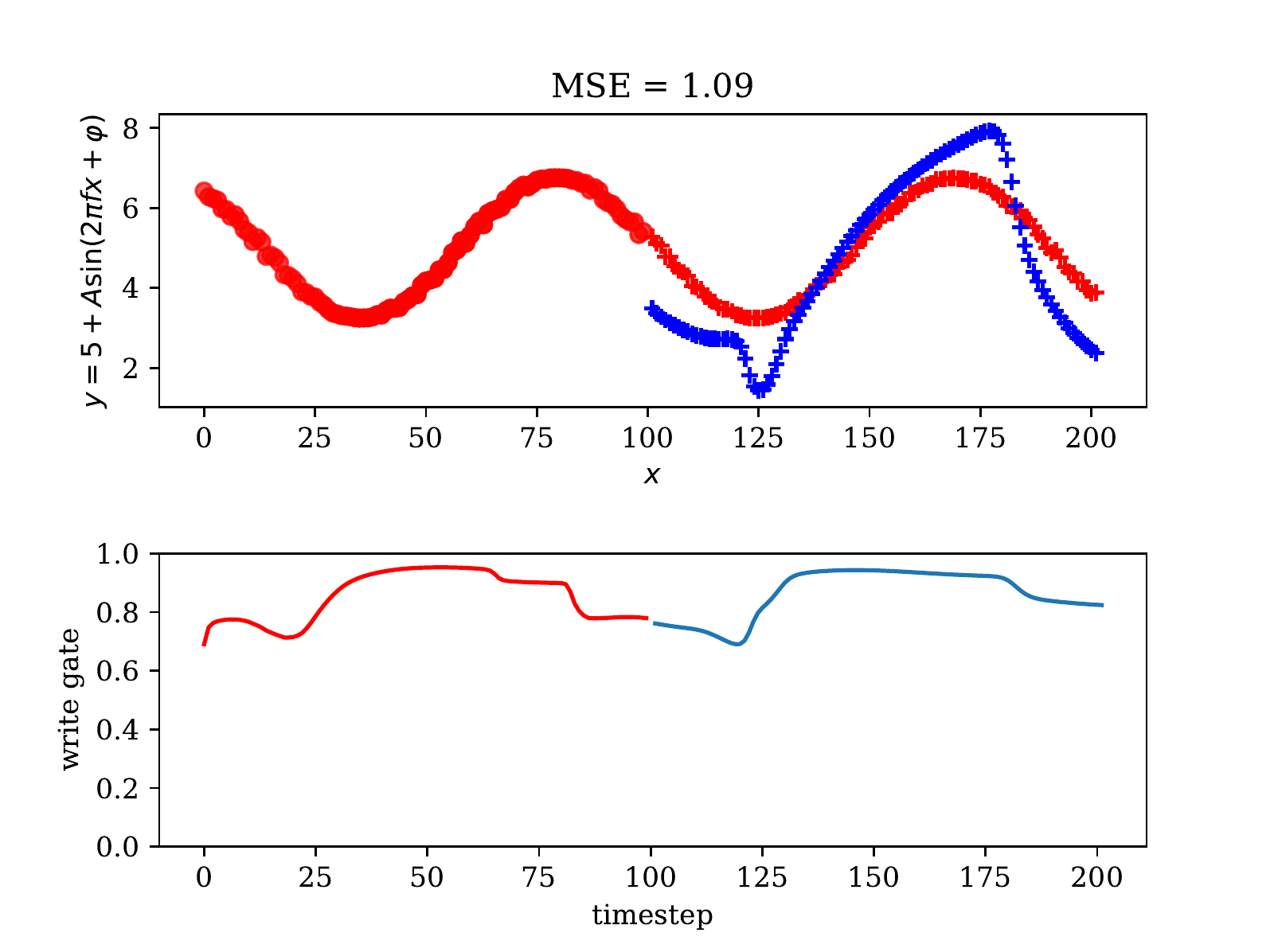}\includegraphics[width=0.3\linewidth]{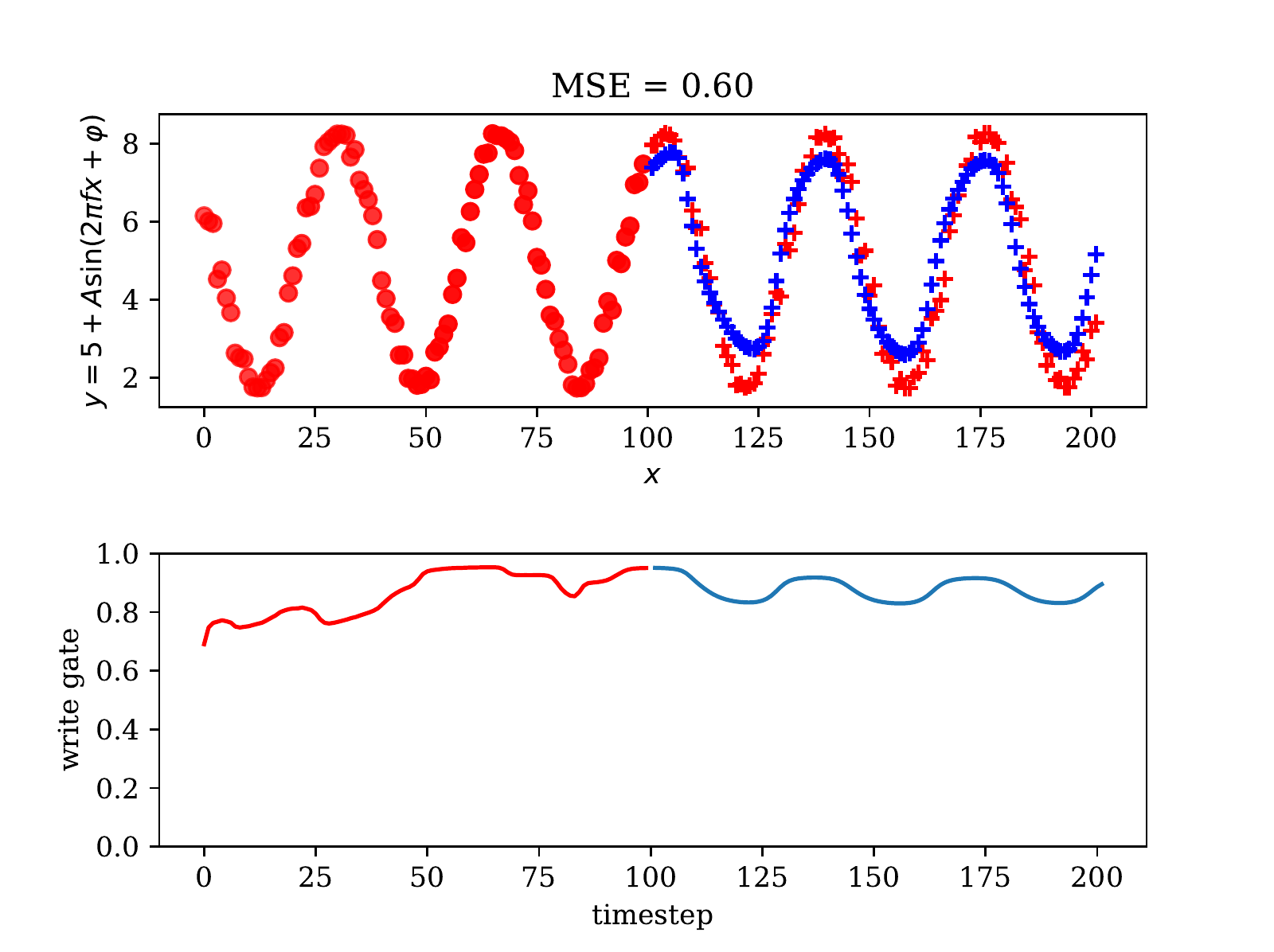}\includegraphics[width=0.3\linewidth]{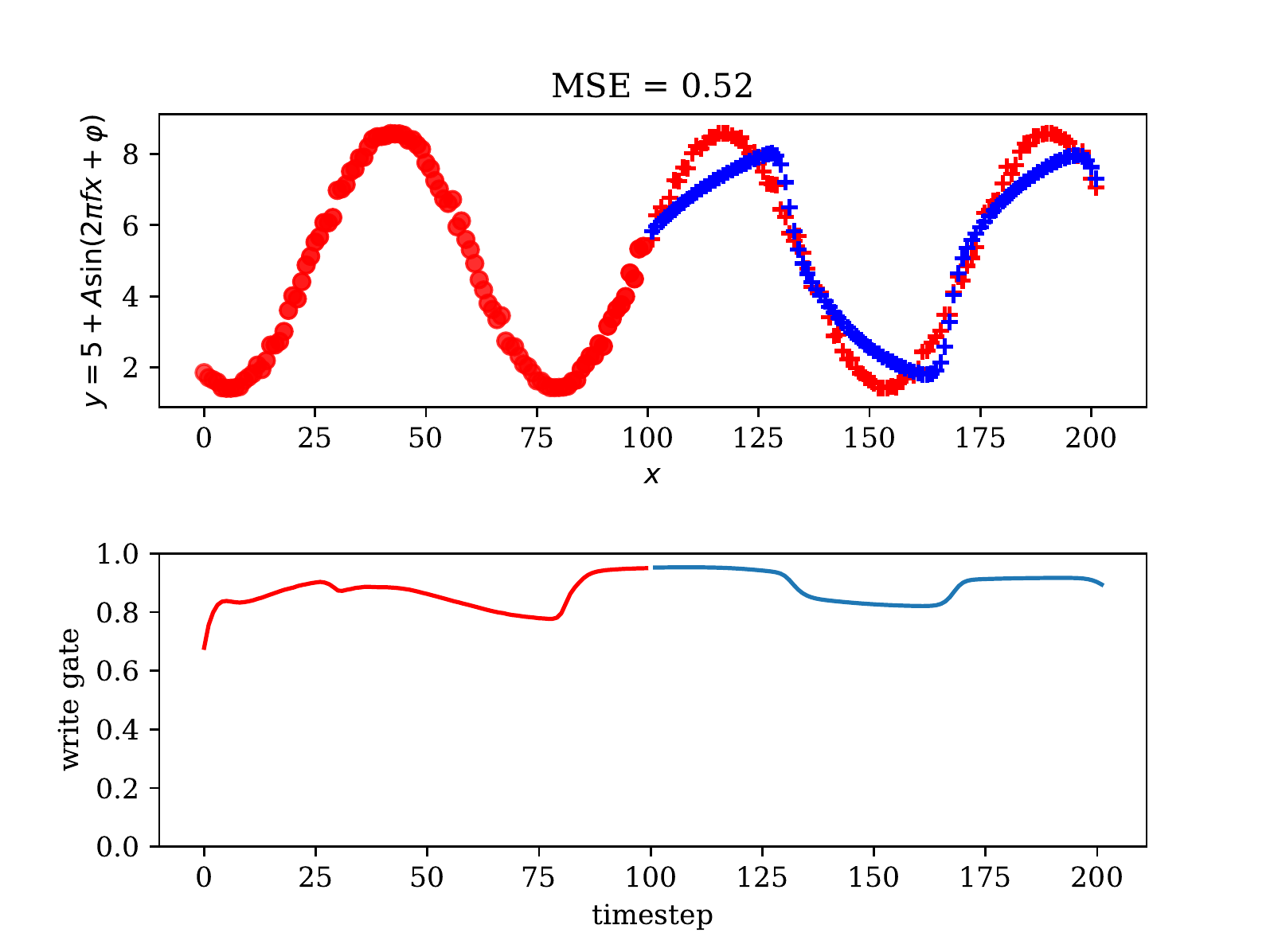}%
\end{minipage}
\par\end{centering}
\begin{centering}
\noindent\begin{minipage}[t]{1\columnwidth}%
\includegraphics[width=0.3\linewidth]{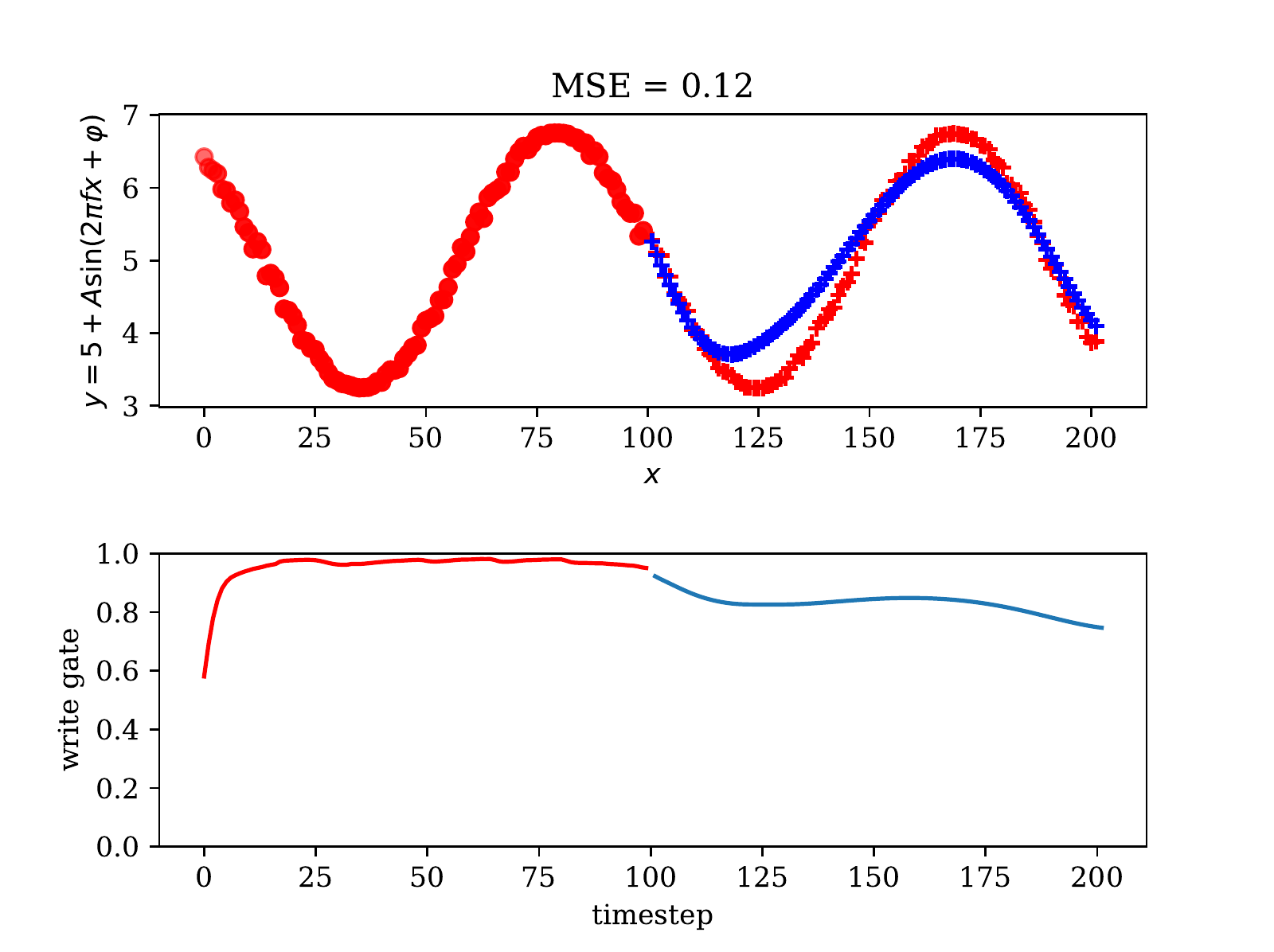}\includegraphics[width=0.3\linewidth]{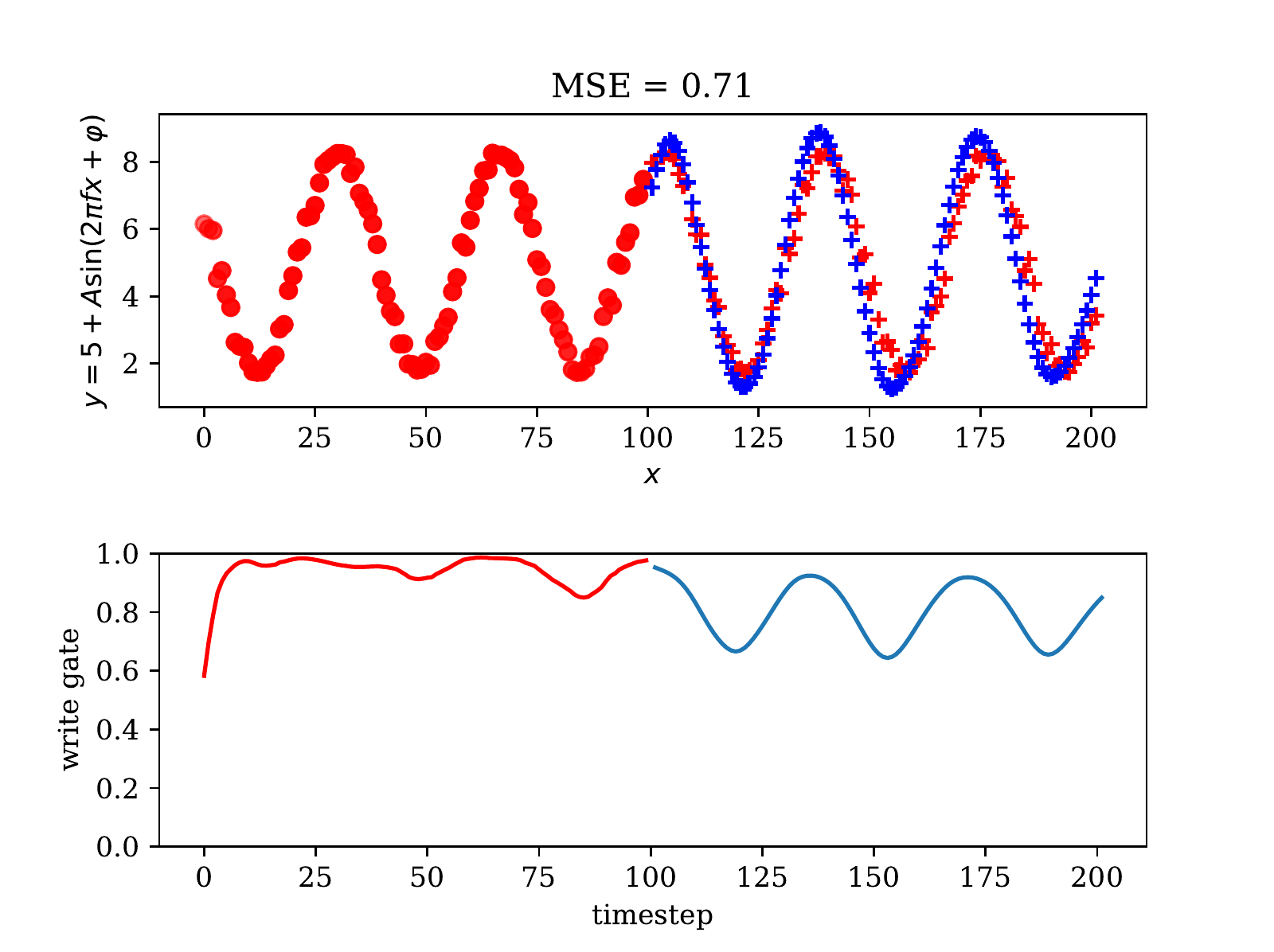}\includegraphics[width=0.3\linewidth]{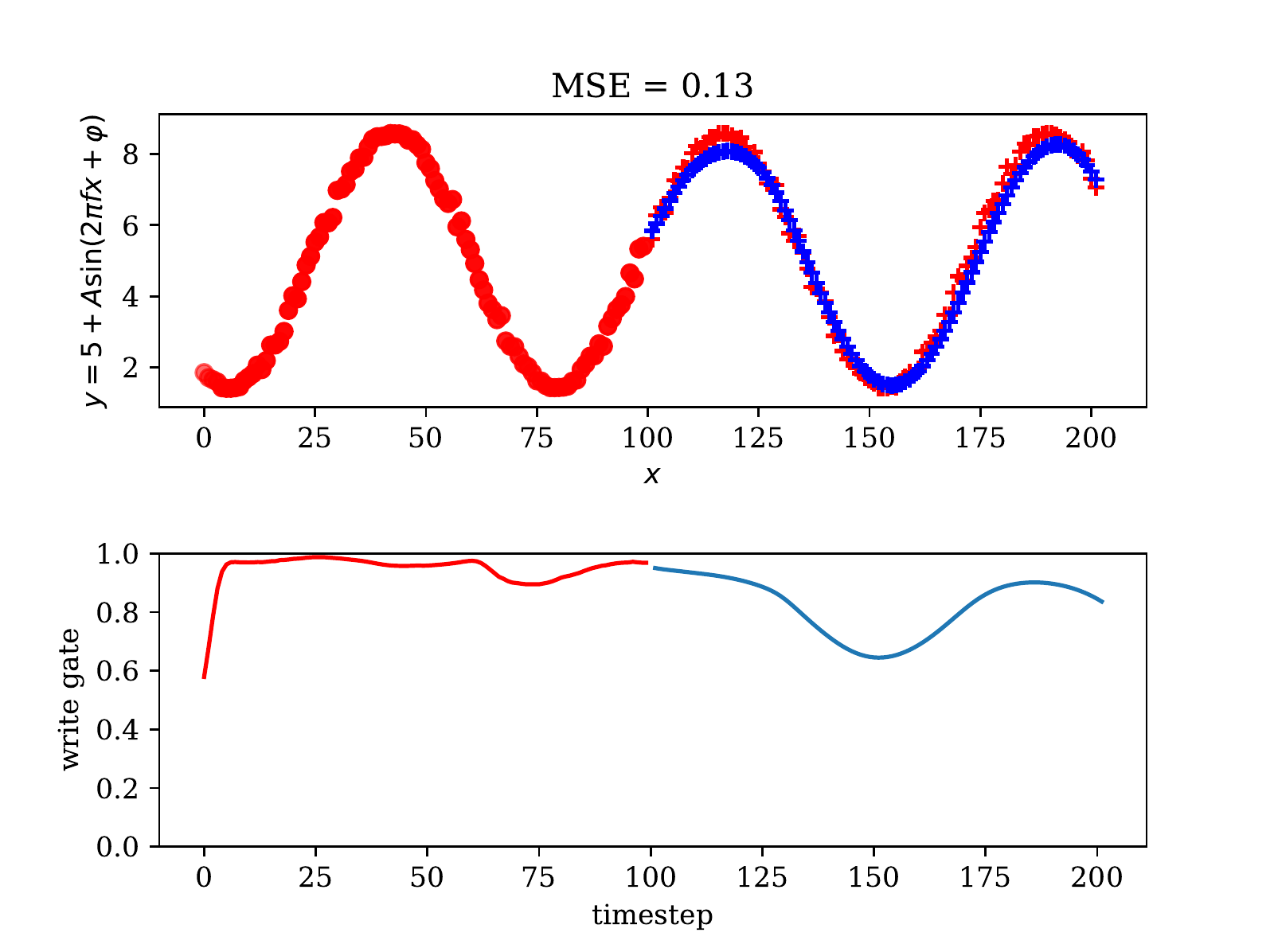}%
\end{minipage}
\par\end{centering}
\begin{centering}
\noindent\begin{minipage}[t]{1\columnwidth}%
\includegraphics[width=0.3\linewidth]{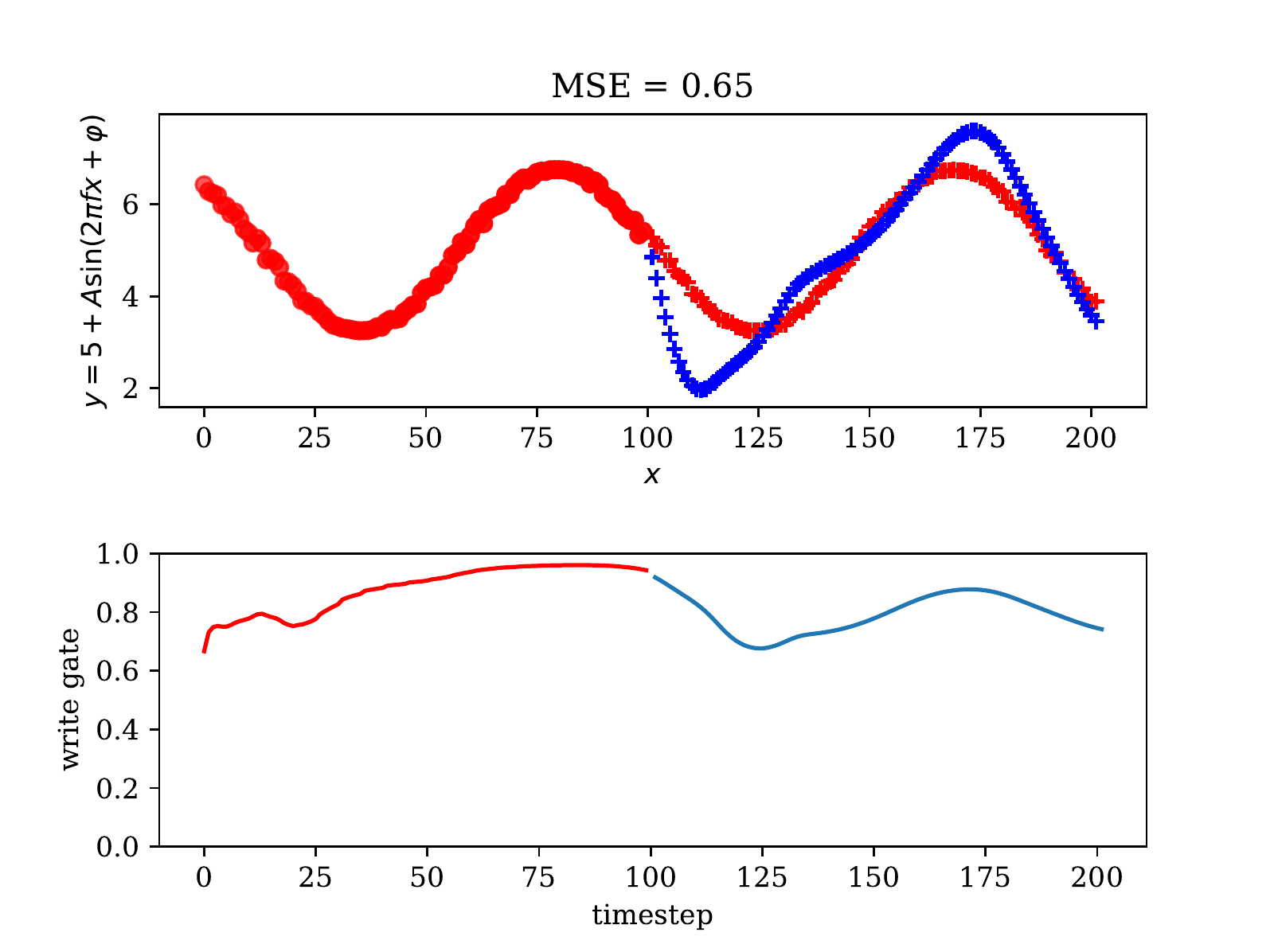}\includegraphics[width=0.3\linewidth]{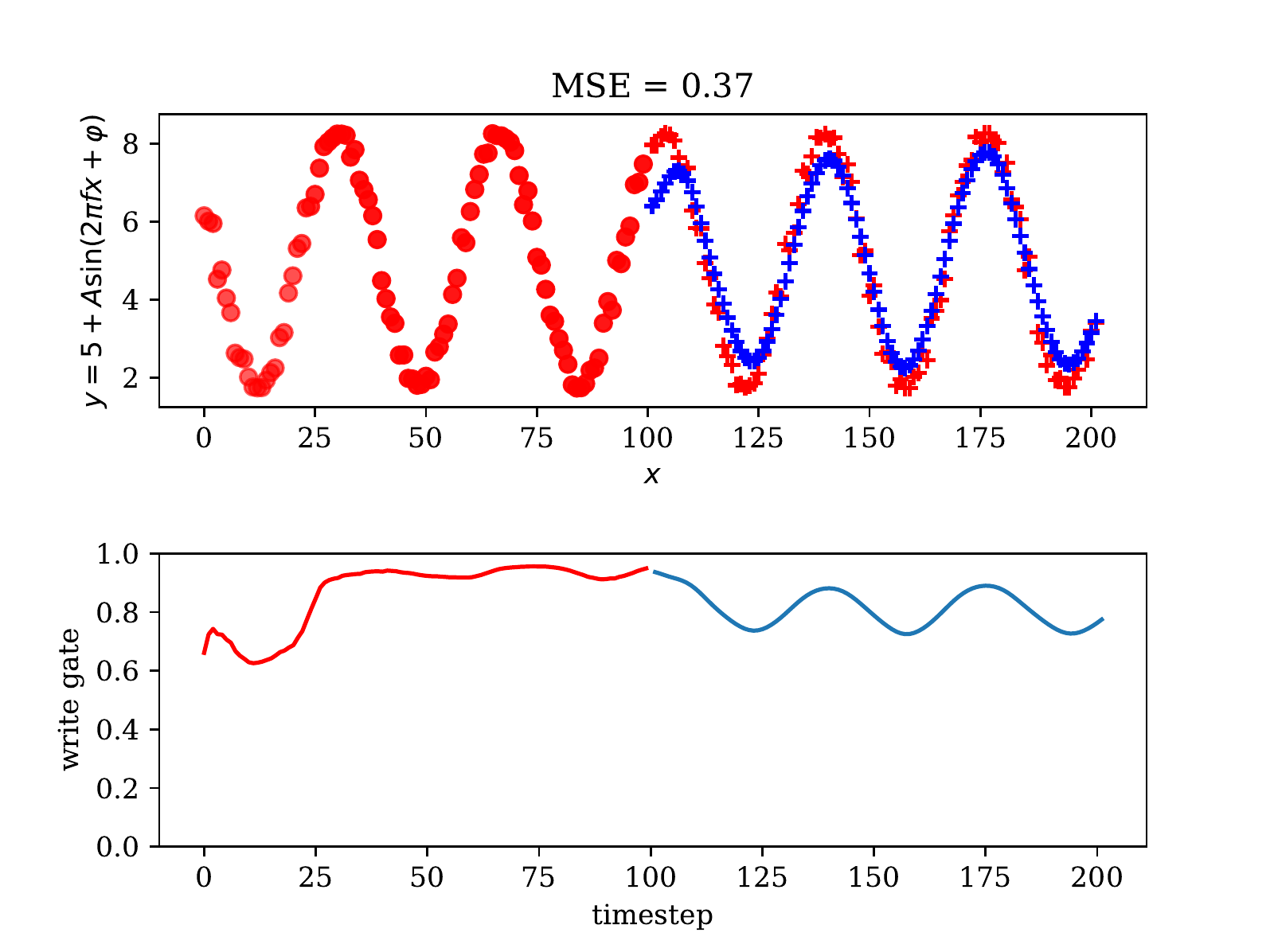}\includegraphics[width=0.3\linewidth]{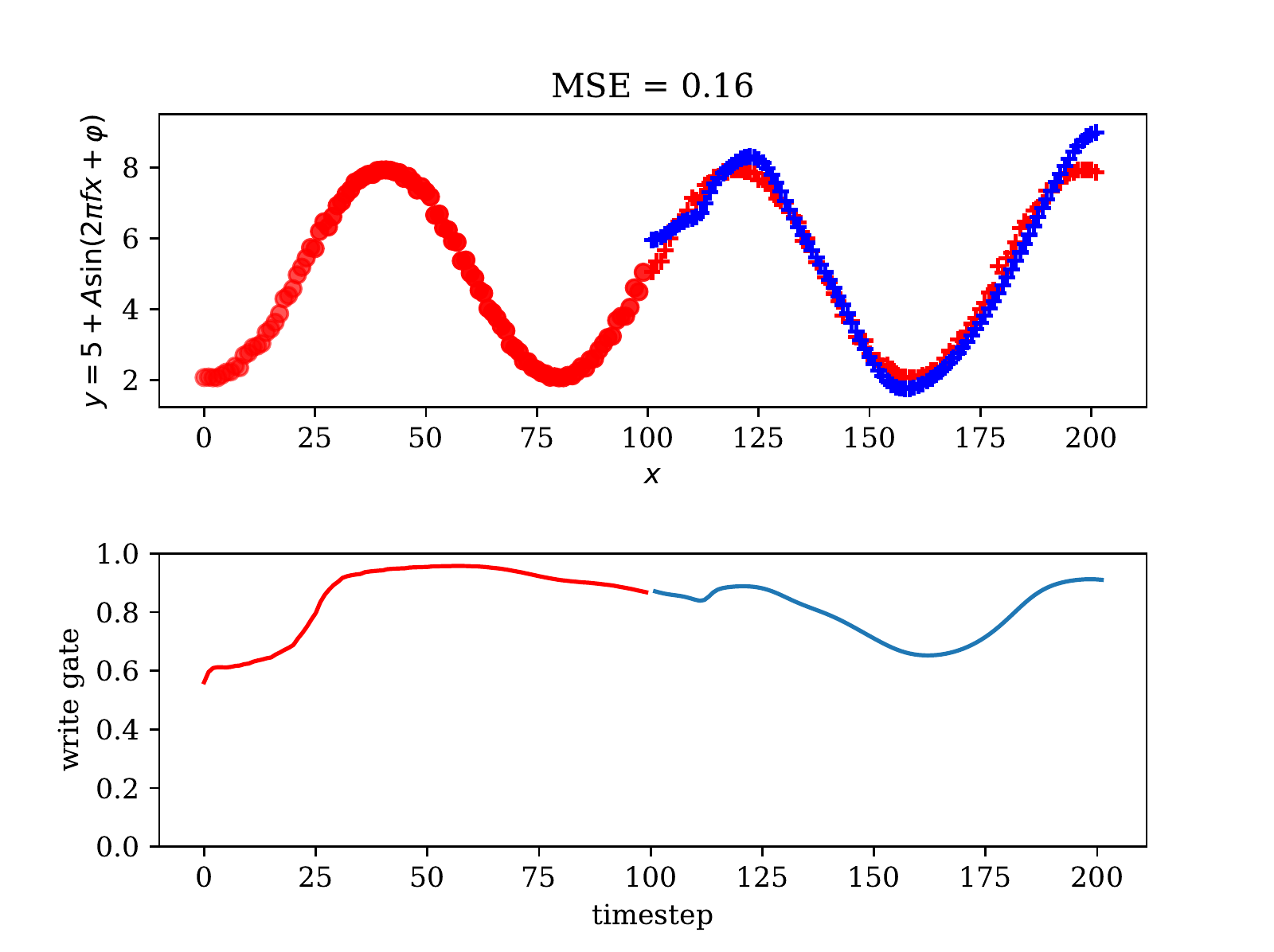}%
\end{minipage}
\par\end{centering}
\caption{Sinusoidal generation with clean input sequence for DNC, UW and CUW
in top-down order. \label{fig:Sinusoid}}
\end{figure}

\begin{figure}
\begin{centering}
\noindent\begin{minipage}[t]{1\columnwidth}%
\includegraphics[width=0.3\linewidth]{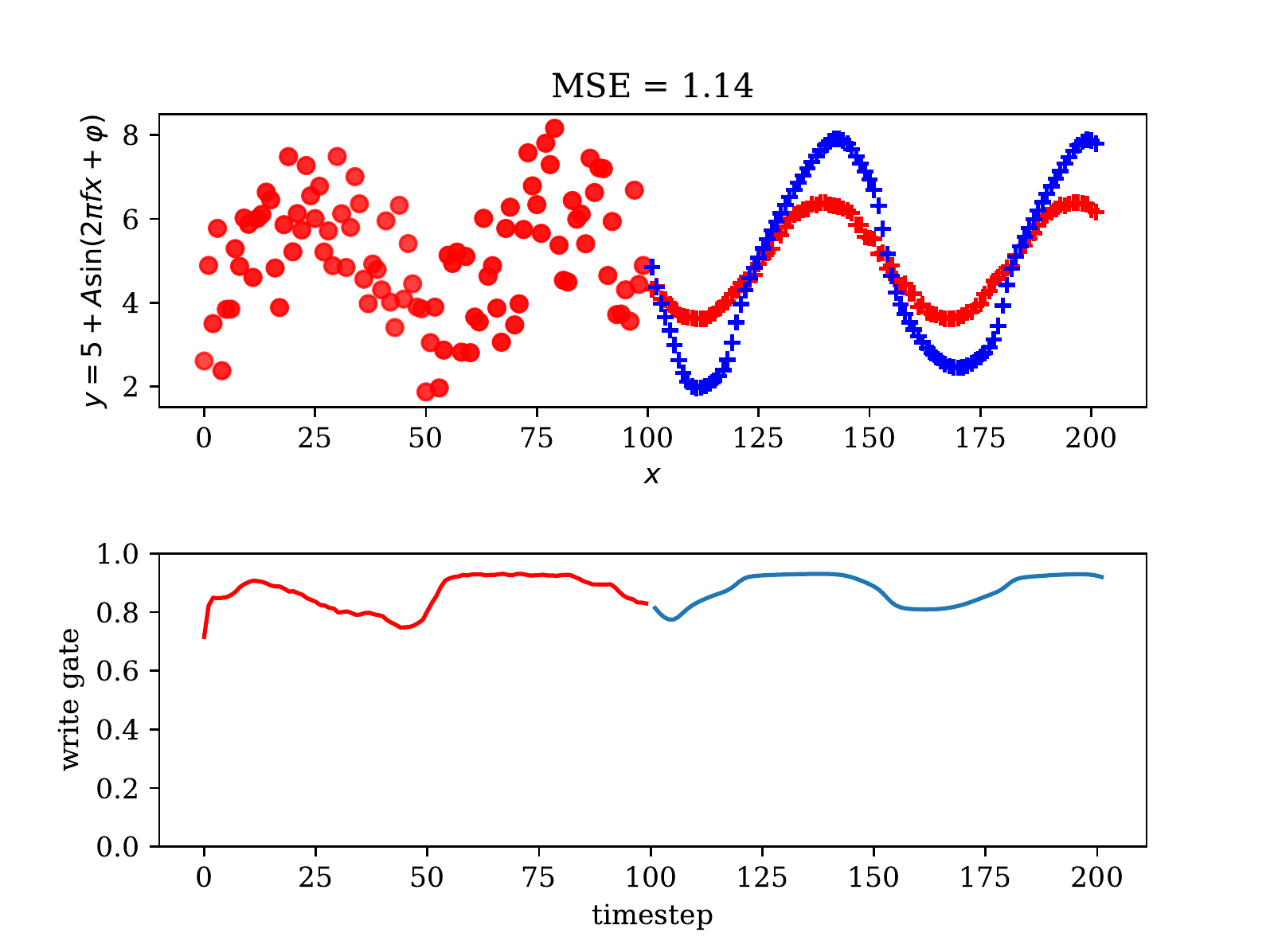}\includegraphics[width=0.3\linewidth]{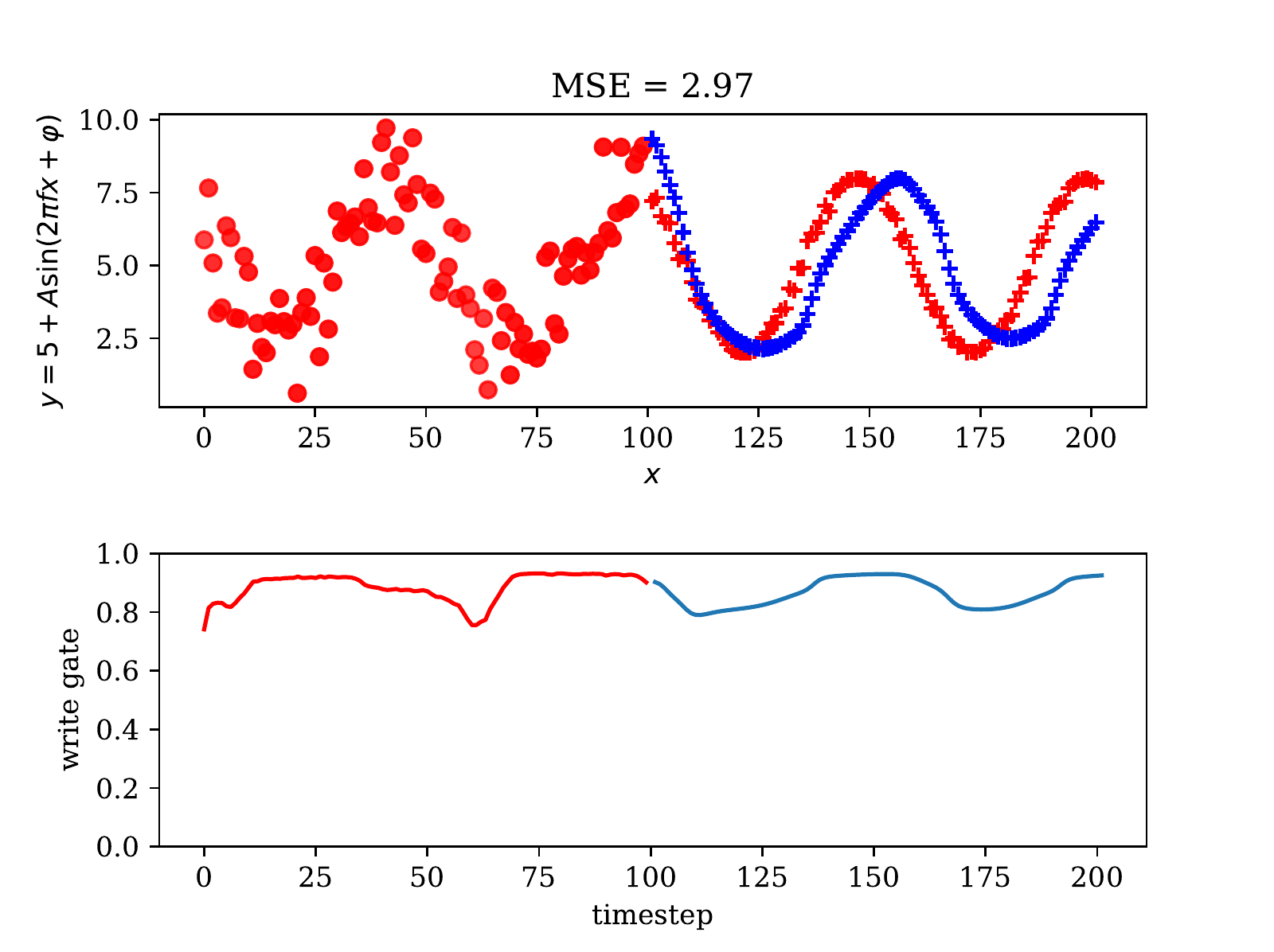}\includegraphics[width=0.3\linewidth]{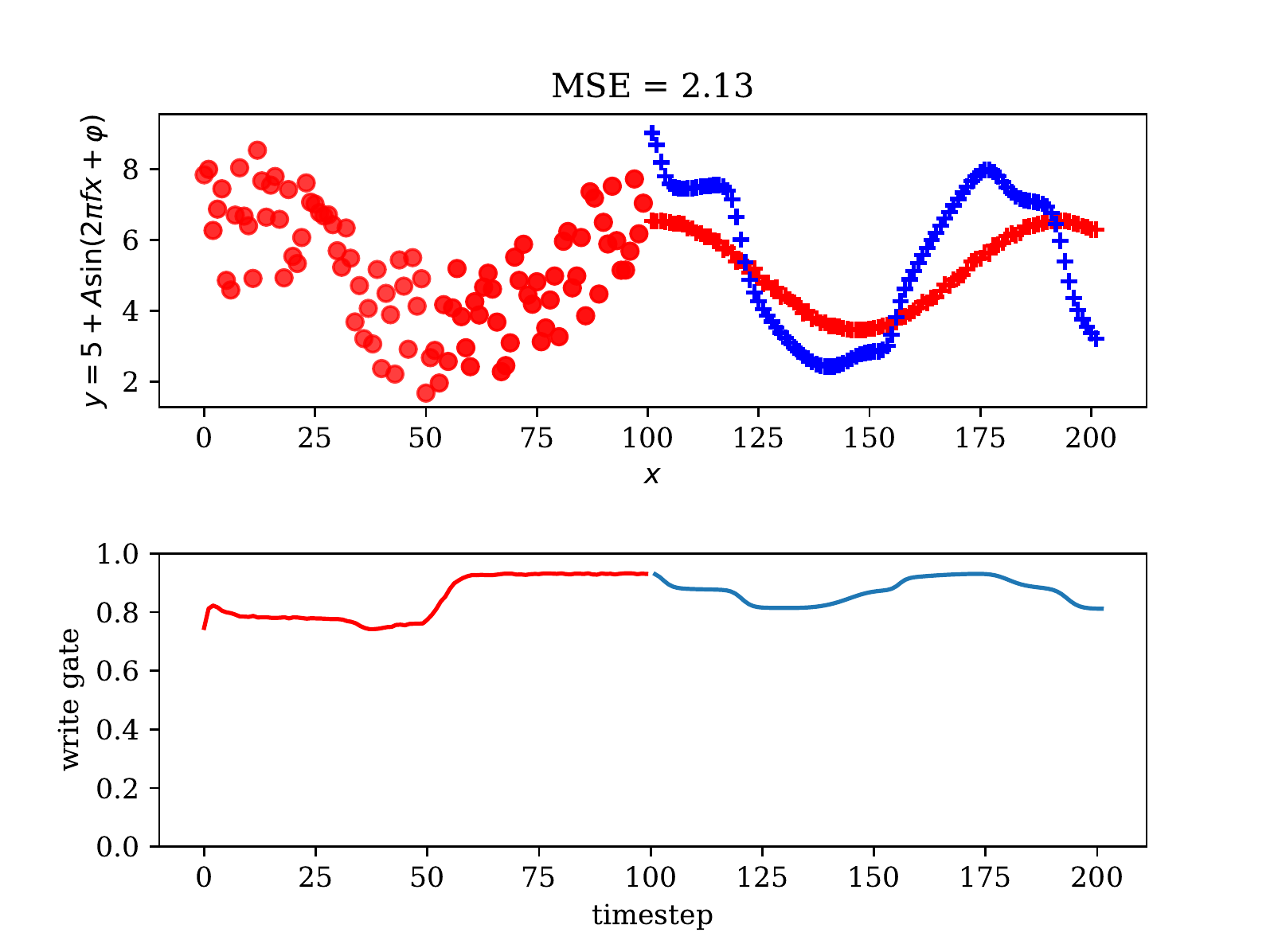}%
\end{minipage}
\par\end{centering}
\begin{centering}
\noindent\begin{minipage}[t]{1\columnwidth}%
\includegraphics[width=0.3\linewidth]{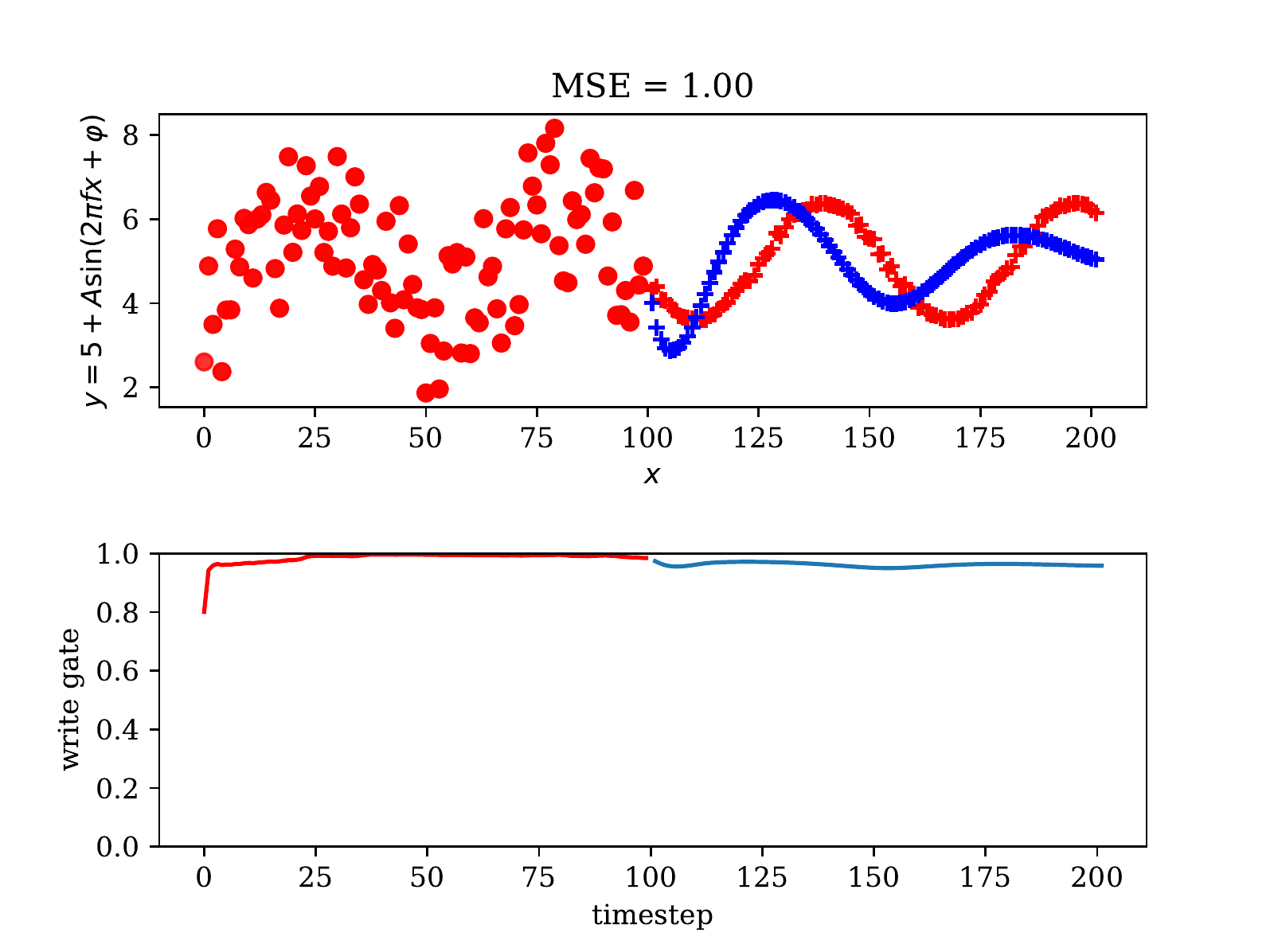}\includegraphics[width=0.3\linewidth]{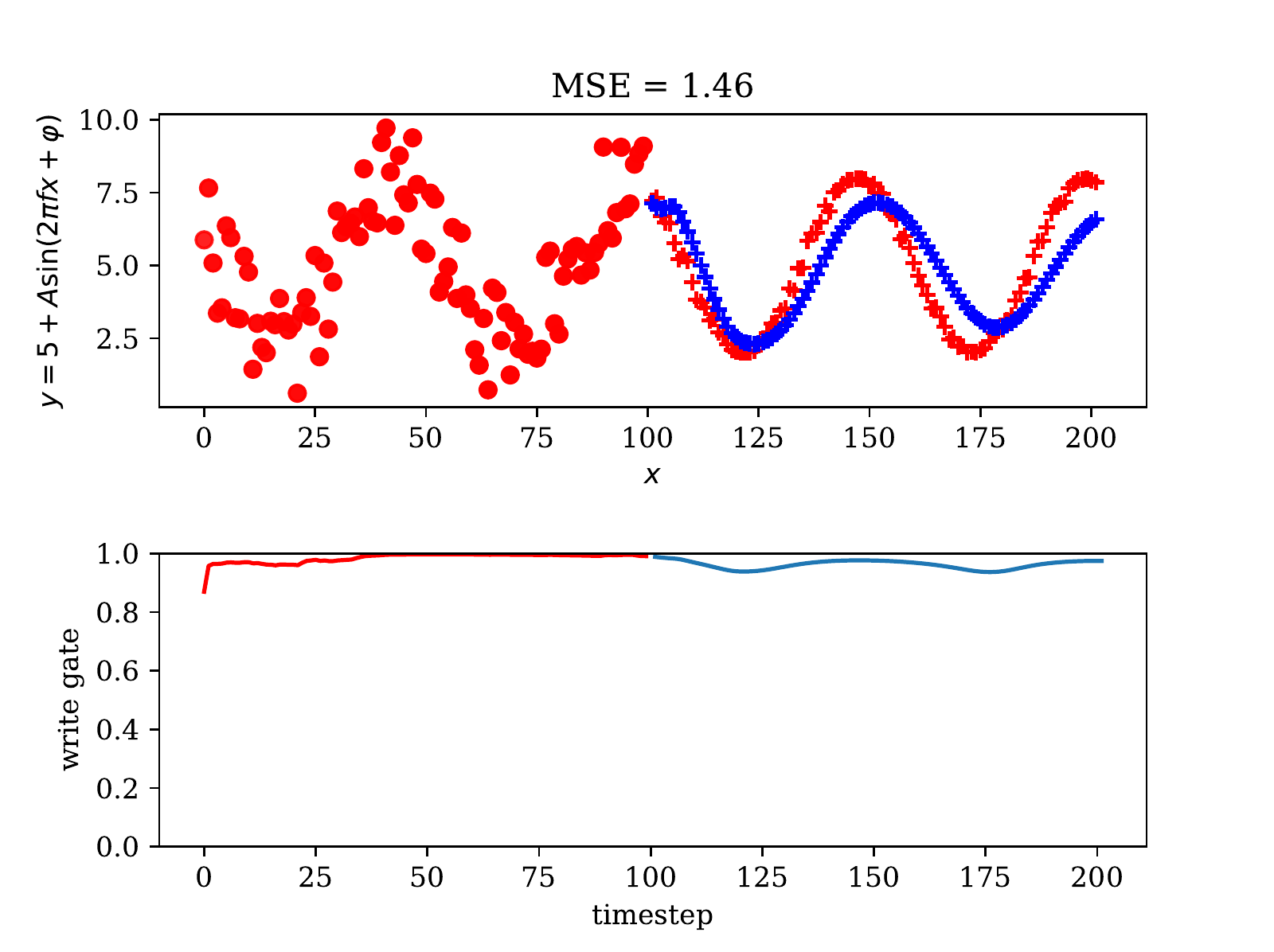}\includegraphics[width=0.3\linewidth]{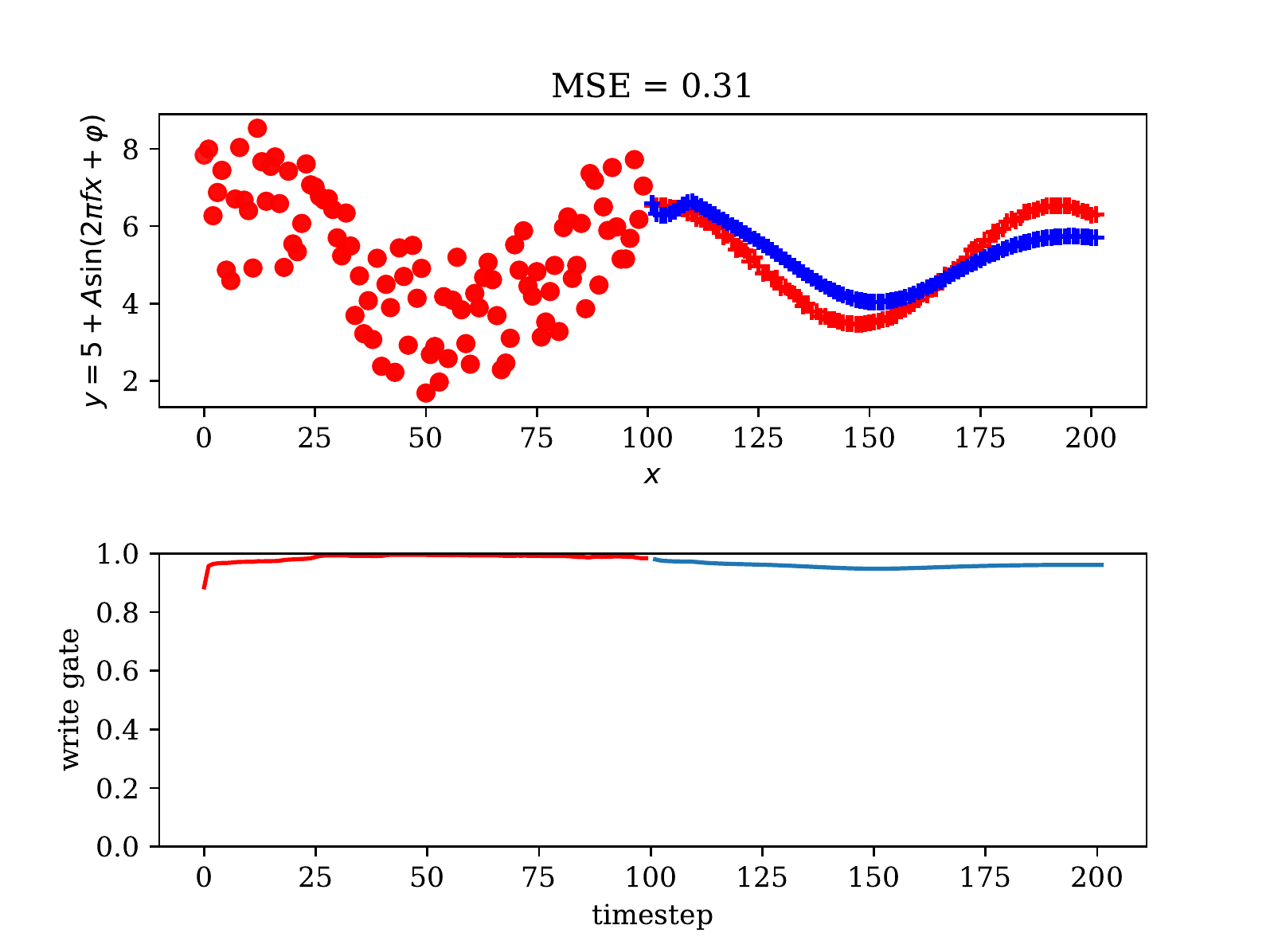}%
\end{minipage}
\par\end{centering}
\begin{centering}
\noindent\begin{minipage}[t]{1\columnwidth}%
\includegraphics[width=0.3\linewidth]{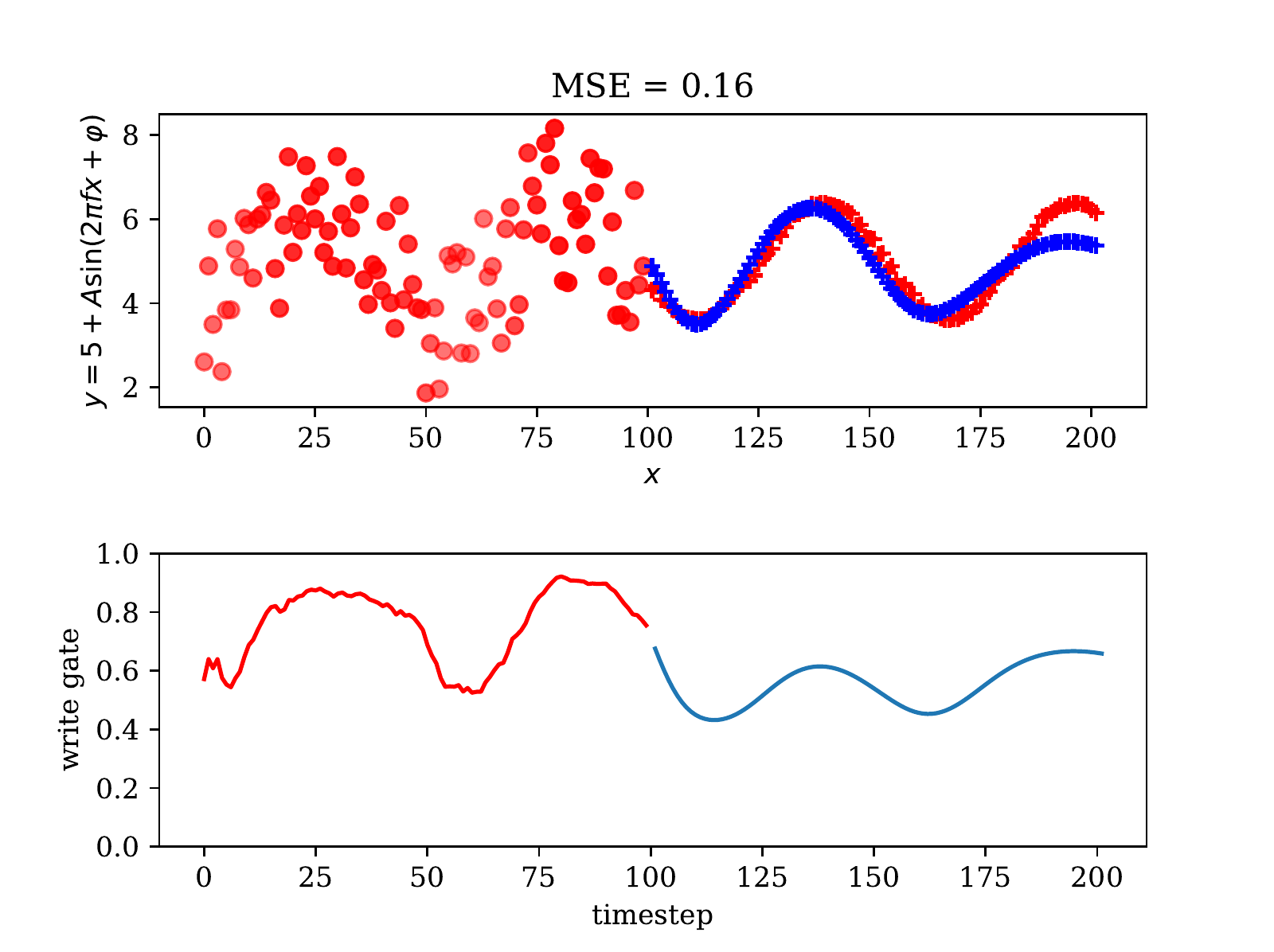}\includegraphics[width=0.3\linewidth]{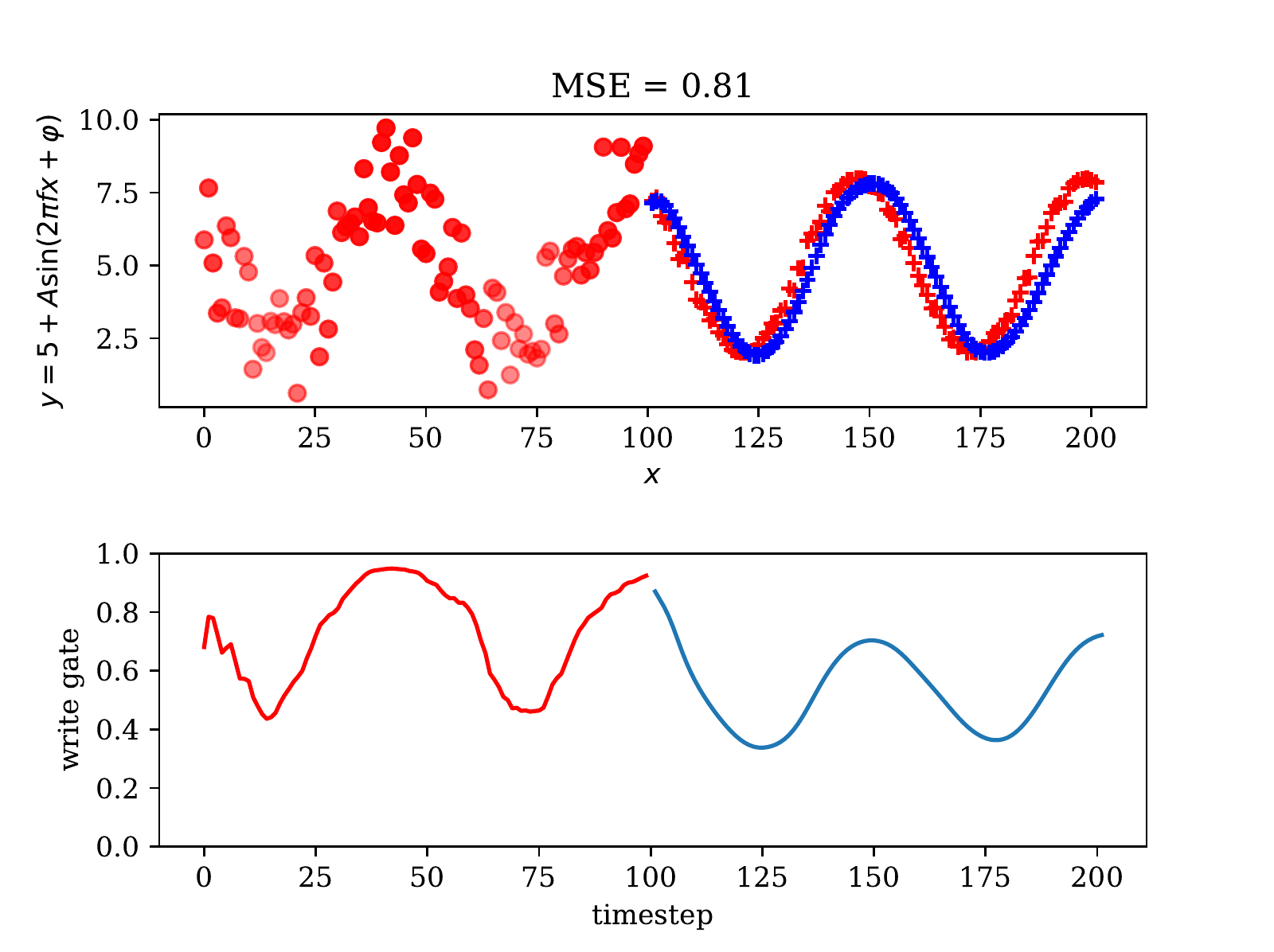}\includegraphics[width=0.3\linewidth]{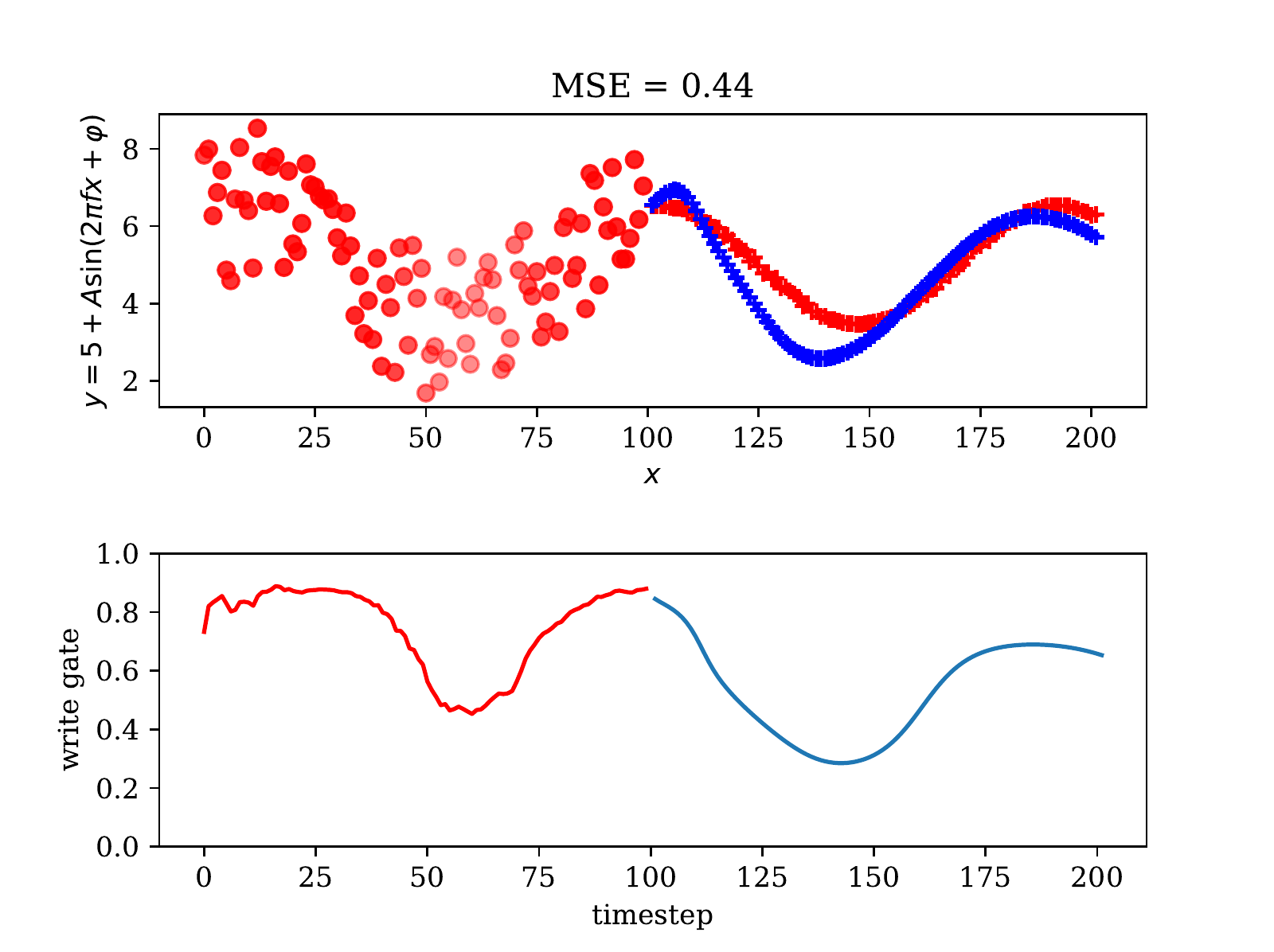}%
\end{minipage}
\par\end{centering}
\caption{Sinusoidal generation with noisy input sequence for DNC, UW and CUW
in top-down order. \label{fig:Sinusoid-noisy}}
\end{figure}

We pick randomly 3 input sequences and plot the output sequences produced
by DNC, UW and CUW in Figs. \ref{fig:Sinusoid} (clean) and \ref{fig:Sinusoid-noisy}
(noisy). In each plot, the first and last 100 timesteps correspond
to the given input and generated output, respectively. The ground
truth sequence is plotted in red while the predicted in blue. We also
visualize the values of MANN write gates through time in the bottom
of each plots. In irregular writing encoding phase, the write gate
is computed even when there is no write as it reflects how much weight
the controller puts on the timesteps. In decoding, we let MANNs write
to memory at every timestep to allow instant update of memory during
inference. 

Under clean condition, all models seem to attend more to late timesteps
during encoding, which makes sense as focusing on late periods of
sine wave is enough for later reconstruction. However, this pattern
is not clear in DNC and UW as in CUW. During decoding, the write gates
tend to oscillate in the shape of sine wave, which is also a good
strategy as this directly reflects the amplitude of generation target.
In this case, both UW and CUW demonstrate this behavior clearer than
DNC. 

Under noisy condition, DNC and CUW try to follow sine-shape writing
strategy. However, only CUW can learn the pattern and assign write
values in accordance with the signal period, which helps CUW decoding
achieve highest accuracy. On the other hand, UW choose to assign write
value equally and relies only on its maximization of timestep contribution.
Although it achieves better results than DNC, it underperforms CUW. 


\subsection{Comparison with non-recurrent methods in flatten image classification
task\label{subsec:Comparsion-with-non-recurrent}}

\begin{table}[H]
\begin{centering}
\begin{tabular}{|l|c|c|}
\hline 
Model & MNIST & pMNIST\tabularnewline
\hline 
\hline 
DNC+CUW & 99.1 & 96.3\tabularnewline
\hline 
The Transformer$^{\star}$ & 98.9 & 97.9\tabularnewline
\hline 
Dilated CNN$^{\blacklozenge}$ & 98.3 & 96.7\tabularnewline
\hline 
\end{tabular}
\par\end{centering}
\caption{Test accuracy (\%) on MNIST, pMNIST. Previously reported results are
from \cite{vaswani2017attention}$^{\star}$ and \cite{chang2017dilated}$^{\blacklozenge}$.
\label{tab:mnist-1}}
\end{table}

\subsection{Details on document classification datasets\label{subsec:Details-on-document}}

\begin{table}[H]
\begin{centering}
\begin{tabular}{|c|c|>{\centering}p{0.1\linewidth}|>{\centering}p{0.1\linewidth}|>{\centering}p{0.1\linewidth}|>{\centering}p{0.1\linewidth}|}
\hline 
Dataset & Classes & Average lengths & Max lengths & Train samples & Test samples\tabularnewline
\hline 
\hline 
IMDb & 2 & 282 & 2,783 & 25,000 & 25,000\tabularnewline
\hline 
Yelp Review Polarity (Yelp P.) & 2 & 156 & 1,381 & 560,000 & 38,000\tabularnewline
\hline 
Yelp Review Full (Yelp F.) & 5 & 158 & 1,381 & 650,000 & 50,000\tabularnewline
\hline 
AG's News (AG) & 4 & 44 & 221 & 120,000 & 7,600\tabularnewline
\hline 
DBPedia (DBP) & 14 & 55 & 1,602 & 560,000 & 70,000\tabularnewline
\hline 
Yahoo! Answers (Yah. A.) & 10 & 112 & 4,392 & 1,400,000 & 60,000\tabularnewline
\hline 
\end{tabular}
\par\end{centering}
\caption{Statistics on several big document classification datasets}

\end{table}

\subsection{Document classification detailed records\label{subsec:Document-classification-detailed}}

\begin{table}[H]
\begin{centering}
\begin{tabular}{|c|c|c|c|c|c|}
\hline 
\multicolumn{2}{|c|}{Model} & AG & IMDb & Yelp P. & Yelp F.\tabularnewline
\hline 
\hline 
\multirow{4}{*}{UW} & 1 & 93.42 & \textbf{91.39} & \textbf{96.39} & 64.89\tabularnewline
\cline{2-6} 
 & 2 & 93.52 & 91.30 & 96.31 & 64.97\tabularnewline
\cline{2-6} 
 & 3 & \textbf{93.69} & 91.25 & 96.39 & \textbf{65.26}\tabularnewline
\cline{2-6} 
 & Mean/Std & 93.54$\pm$0.08 & 91.32$\pm$0.04 & 96.36$\pm$0.03 & 65.04$\pm$0.11\tabularnewline
\hline 
\multirow{4}{*}{CUW} & 1 & 93.61 & 91.26 & \textbf{96.42} & \textbf{65.63}\tabularnewline
\cline{2-6} 
 & 2 & \textbf{93.87} & 91.18 & 96.29 & 65.05\tabularnewline
\cline{2-6} 
 & 3 & 93.70 & \textbf{91.32} & 96.36 & 64.80\tabularnewline
\cline{2-6} 
 & Mean/Std & 93.73$\pm$0.08 & 91.25$\pm$0.04 & 96.36$\pm$0.04 & 65.16$\pm$0.24\tabularnewline
\hline 
\end{tabular}
\par\end{centering}
\caption{Document classification accuracy (\%) on several datasets reported
for 3 different runs. Bold denotes the best records.}

\end{table}